\documentclass{article}

    \usepackage[final]{neurips_2024}

\usepackage[utf8]{inputenc} 
\usepackage[T1]{fontenc}    
\usepackage{hyperref}       
\usepackage{url}            
\usepackage{booktabs}       
\usepackage{amsfonts}       
\usepackage{nicefrac}       
\usepackage{microtype}      
\usepackage{xcolor}         
\usepackage{latexsym}
\usepackage{amssymb}
\usepackage{caption}
\usepackage{subcaption}
\usepackage{amsmath}
\usepackage{amsthm}
\usepackage{enumitem}
\usepackage{color}
\usepackage{algorithm}
\usepackage{subcaption}
\usepackage{graphicx}
\usepackage{algorithmic}
\usepackage{multirow}
\usepackage{wrapfig}
\usepackage{siunitx} 
\hypersetup{hidelinks,
    colorlinks=true,
    anchorcolor=blue,
    citecolor={olive},
    linkcolor=blue}

\title{

Diffusion Actor-Critic with Entropy Regulator}

\author{%
Yinuo Wang$^1$ \quad Likun Wang$^1$ \quad Yuxuan Jiang$^1$ \quad Wenjun Zou$^1$ \quad Tong Liu$^1$ \\
\textbf{Xujie Song$^1$ \quad Wenxuan Wang$^1$ \quad Liming Xiao$^{2}$ \quad Jiang Wu$^{2}$} \\
\textbf{Jingliang Duan$^{1,2*}$ \quad Shengbo Eben Li$^{1}\thanks{Corresponding author <duanjl15@163.com> <lishbo@tsinghua.edu.cn>.}$} \vspace{0.15cm}\\
$^1$School of Vehicle and Mobility, Tsinghua University \\ $^2$School of Mechanical Engineering, University of Science and Technology Beijing
}

\begin{document}

\maketitle

\begin{abstract}
Reinforcement learning (RL) has proven highly effective in addressing complex decision-making and control tasks. However, in most traditional RL algorithms, the policy is typically parameterized as a diagonal Gaussian distribution with learned mean and variance, which constrains their capability to acquire complex policies. In response to this problem, we propose an online RL algorithm termed diffusion actor-critic with entropy regulator (DACER). This algorithm conceptualizes the reverse process of the diffusion model as a novel policy function and leverages the capability of the diffusion model to fit multimodal distributions, thereby enhancing the representational capacity of the policy. Since the distribution of the diffusion policy lacks an analytical expression, its entropy cannot be determined analytically. To mitigate this, we propose a method to estimate the entropy of the diffusion policy utilizing Gaussian mixture model. Building on the estimated entropy, we can learn a parameter $\alpha$ that modulates the degree of exploration and exploitation. Parameter $\alpha$ will be employed to adaptively regulate the variance of the added noise, which is applied to the action output by the diffusion model. Experimental trials on MuJoCo benchmarks and a multimodal task demonstrate that the DACER algorithm achieves state-of-the-art (SOTA) performance in most MuJoCo control tasks while exhibiting a stronger representational capacity of the diffusion policy.
\end{abstract}

\section{Introduction}
Recently, deep reinforcement learning (RL) has emerged as an effective method for solving optimal control problems in the physical world \cite{guan2021integrated, peng2021end, kaufmann2023champion, li2023reinforcement}. In most existing RL algorithms, the policy is parameterized as a deterministic function or a diagonal Gaussian distribution with the learned mean and variance \cite{schulman2015trust, schulman2017proximal, haarnoja2018soft, duan2021distributional}. However, the theoretically optimal policy may exhibit strong multimodality, which cannot be well modeled by deterministic or diagonal Gaussian policies \cite{wang2022diffusion, kang2024efficient, yang2023policy}. Restricted policy representation capabilities can make algorithms prone to local optimal solutions, damaging policy performance. For instance, in situations where two distinct actions in the same state yield approximately the same Q-value, the Gaussian policy approximates the bimodal action by maximizing the Q-value. This results in the policy displaying mode-covering behavior, concentrating high density in the intermediate region between the two patterns, which is inherently a low-density region with a lower Q-value. Consequently, modeling the policy with a unimodal Gaussian distribution is likely to significantly impair policy learning.

Lately, the diffusion model has become widely known as a generative model for its powerful ability to fit multimodal distributions  \cite{ho2020denoising, song2021maximum, croitoru2023diffusion}. It learns the original data distribution through the idea of stepwise addition and removal of noise and has excellent performance in the fields of image \cite{yang2023diffusion, peebles2023scalable} and video generation \cite{dhariwal2021diffusion, blattmann2023stable}. The policy network in RL can be seen as a state-conditional generative model. Given the ability of diffusion models to fit complex distributions, there is increasing work on combining RL with diffusion models. Online RL learns policies by interacting with the environment \cite{haarnoja2018soft, schulman2017proximal}. Offline RL, also known as batch RL, aims to effectively learn policies from previously collected data without interacting with the environment \cite{agarwal2020optimistic, brandfonbrener2021offline}. In practical applications, many control problems have excellent simulators. At this time, using offline RL is not appropriate, as online RL with interaction capabilities performs better. Therefore, this paper focuses on how the diffusion model can be combined with online RL.



In this work, we propose diffusion actor-critic with entropy regulator (DACER), a generalized new approach to combine diffusion policy with online RL. Specifically, we base DACER on the denoising diffusion probabilistic model (DDPM) \cite{ho2020denoising}. A recent work by He \textit{et al.} \cite{he2024diffusion} points out that the representational power of diffusion models stems mainly from the reverse diffusion processes, not from the forward diffusion processes. Inspired by this work, we reconceptualize the reverse process of the diffusion model as a novel policy approximator, leveraging its powerful representation capabilities to enhance the performance of RL algorithms. The optimization objective of this novel policy function is to maximize the expected Q-value. Maximizing entropy is important for policy exploration in RL, but the entropy of the diffusion policy is difficult to determine. Therefore, we choose to sample actions at fixed intervals and use a Gaussian mixture model (GMM) to fit the action distributions. Subsequently, We can calculate the approximate entropy of the policy in each state. The average of these entropies is then used as an approximation of the current diffusion policy entropy. Then, we use the estimated entropy to regulate the degree of exploration and exploitation of diffusion policy.

In summary, the key contributions of this paper are the following: 1) We propose to consider the reverse process of the diffusion model as a novel policy function. The objective function of the diffusion policy is to maximize the expected Q-value and thus achieve policy improvement. 2) We propose a method for estimating the entropy of diffusion policy. The estimated value is utilized to achieve an adaptive adjustment of the exploration level of the diffusion policy, thus improving the policy performance. 3) We evaluate the efficiency and generality of our method on the popular MuJoCo benchmarking. Compared with DDPG \cite{silver2014deterministic}, TD3 \cite{fujimoto2018addressing}, PPO \cite{schulman2017proximal}, SAC \cite{haarnoja2018soft}, DSAC \cite{duan2023dsac, duan2021distributional}, and TRPO \cite{schulman2015trust}, our approach achieves the SOTA performance. In addition, we demonstrate the superior representational capacity of our algorithm through a specific multi-goal task. 4) We provide the DACER code written in JAX to facilitate future researchers to follow our work \footnote[1]{https://github.com/happy-yan/DACER-Diffusion-with-Online-RL}.

Section \ref{sec:related work} introduces and summarizes existing approaches to diffusion policy in offline RL and online RL, pointing out some of their problems. Section \ref{sec:preliminaries} provides an introduction to online RL and diffusion models. Our approach to combining diffusion policy with the mainstream actor-critic framework, as well as methods to enhance the performance of diffusion policy will be presented in section \ref{sec:method}. The results of the experiments in the MuJoCo environment, the ablation experiments as well as the multimodality task will be presented in section \ref{sec:experiment}. Section \ref{sec:conclusion} provides the conclusions of this paper.

\section{Related Work}
\label{sec:related work}
\paragraph{Diffusion Policy in Offline RL}
Offline RL leverages pre-collected datasets for policy development, circumventing direct environmental interaction. Current offline RL research utilizing diffusion models as policy networks primarily adhere to the behavioral cloning framework \cite{codevilla2019exploring, ly2020learning}.  Within this framework, two main objectives emerge: performance enhancement and training efficiency improvement. For the former, Cheng \textit{et al.} \cite{chi2023diffusion} proposed a Diffusion Policy, casting the policy as a conditional denoising diffusion process within the action space to accommodate complex multimodal action distributions. Wang \textit{et al.} \cite{wang2022diffusion} introduced Diffusion-QL, which integrates behavior cloning via diffusion model loss with Q-learning for policy improvement. Ajay \textit{et al.} \cite{ajay2022conditional} created Decision Diffusion, incorporating classifier-free guidance into the diffusion model to integrate trajectory information, such as rewards and constraints. Addressing the latter, Kang \textit{et al.} \cite{kang2023efficient} developed efficient diffusion policies (EDP), an evolution of Diffusion-QL. EDP accelerates training by utilizing initial actions from state-action pairs in the buffer and applying a one-step sample for final action derivation. Chen \textit{et al.} \cite{chen2023boosting} proposed a consistency policy that enhances diffusion algorithm efficiency through one-step action generation from noise during training and inference. Although Diffusion Policy's powerful ability to fit multimodal policy distributions can achieve good performance in offline RL tasks, this method of policy improvement based on behavioral cloning cannot be directly transferred to online RL. In addition, the biggest challenge facing offline RL, the distribution shift problem, has not been completely solved. This paper focuses on online RL, moving away from the framework of behavioral cloning.


\paragraph{Diffusion Policy in Online RL} Online RL, characterized by real-time environment interaction, contrasts with offline RL's dependence on pre-existing datasets. To date, only two studies have delved into integrating online RL with diffusion models. Yang \textit{et al.} \cite{yang2023policy} pioneered this approach by proposing action gradient method. This approach achieves policy improvement by updating the action in the replay buffer through the $\nabla_a Q
$, followed by mimicry learning of the action post-update using a diffusion model. However, action gradient increased the additional training time. Furthermore, it is difficult to fully learn both the action gradient and imitation learning steps simultaneously, which also resulted in suboptimal performance of this method in MuJoCo tasks. Psenka \textit{et al.} \cite{psenka2023learning}  proposed Q-score matching (QSM), a new methodology for off-policy reinforcement learning that leverages the score-based structure of diffusion model policies to align with the $\nabla_a Q
$. This approach aims to overcome the limitations of simple behavior cloning in actor-critic settings by integrating the policy's score with the Q-function's action gradient. However, QSM needs to accurately learn  $\nabla_a Q
$ in most of the action space to achieve optimal guidance. This is difficult to accomplish, resulting in suboptimal performance of QSM. 

Our method is motivated to propose a diffusion policy that can be combined with most existing actor-critic frameworks. We first consider the reverse diffusion process of the diffusion model as a policy function with strong representational power. Then, we use the entropy estimation method to balance the exploration and utilization of diffusion policy and improve the performance of the policy.

\paragraph{Comparison with Diffusion-QL} Diffusion-QL \cite{wang2022diffusion} made a successful attempt by replacing the diagonal Gaussian policy with a diffusion model. It also guides the updating of the policy by adding the normalized Q-value in the policy loss term. The main differences between our work and Diffusion-QL are as follows: 1) Diffusion-QL is still essentially an architecture for imitation learning, and policy updates are mainly motivated by the imitation learning loss term. 2) Our work adaptively regulates the standard deviation of random noise in the sampling process $\boldsymbol{a} = \boldsymbol{a} + \lambda \alpha \cdot \mathcal{N}(0, \boldsymbol{I})$, where $\alpha$ is a learned parameter, $\lambda$ is a hyperparameter. This method effectively balances exploration and exploitation and subsequently enhances the performance of the diffusion policy. Ablation experiments provide evidence supporting these findings.
\section{Preliminaries}
\label{sec:preliminaries}

\subsection{Online Reinforcement Learning}
\label{sec:pre_onlineRL}
In the conventional framework of RL, interactions between the agent and its environment occur in sequential discrete time steps. Typically, the environment is modeled as a Markov decision process (MDP) with continuous states and actions \cite{sutton2018reinforcement}. The environment provides feedback through a bounded reward function denoted by $r(s_t, a_t)$. The likelihood of transitioning to a new state based on the agent's action is expressed by the probability $p(s_{t+1} | s_t, a_t)$. State-action pairs for the current and next steps are indicated as $(s, a)$ and $(s', a')$. The decision-making of an agent at any state $s_t$ is guided by a stochastic policy $\pi(a_t|s_t)$, which determines the probability distribution over feasible actions at that state.

In the realm of online RL, agents engage in real-time learning and decision-making through direct interactions with their environments. Such interactions are captured within a tuple $(s_t,a_t,r_t,s_{t+1})$, representing the transition during each interaction. It is common practice to store these transitions in an experience replay buffer, symbolized as $\mathcal{B}$. Throughout the training phase, random samples drawn from $\mathcal{B}$ produce batches of data that contribute to a more consistent training process. The fundamental aim of traditional online RL strategies is to craft a policy that optimizes the expected total reward:
\begin{equation}
J_\pi=\mathbb{E}_{(s_{i\geq t},a_{i\geq t})\sim\pi}\Big[\sum_{i=t}^\infty\gamma^{i-t}r(s_i, a_i)\Big],
\end{equation}
where $\gamma\in(0,1)$ represents the discount factor. The Q-value for a state-action pair $(s,a)$ is given by
\begin{equation}
Q(s,a)=\mathbb{E} _ \pi\Big[\sum _ {i=0}^\infty\gamma^ir(s_i, a_i)|s_0=s,a_0=a\Big].
\end{equation}

RL typically employs an actor-critic framework \cite{li2023reinforcement, konda1999actor}, which includes both a policy function, symbolized by $\pi$, and a corresponding Q-value function, noted as $Q^{\pi}$. The process of policy iteration is often used to achieve the optimal policy $\pi^*$, cycling through phases of policy evaluation and enhancement. In the policy evaluation phase, the Q-value $Q^{\pi}$ is recalibrated according to the self-consistency requirements dictated by the Bellman equation:
\begin{equation}
  \label{eq:bellman}
    Q^{\pi}(s,a) = r(s,a) + \gamma \mathbb{E}_{s' \sim p, a' \sim \pi}[Q^{\pi}(s',a')].
\end{equation}

In the policy improvement phase, an enhanced policy $\pi_{\mathrm{new}}$ is sought by optimizing current Q-value  $Q^{\pi_{\mathrm{old}}}$:
\begin{equation}
  \label{eq:policy}
  \pi_{\mathrm{new}} = \arg \max_{\pi}  \mathbb{E}_{s \sim d_{\pi},a \sim \pi }[Q^{\pi_{\mathrm{old}}}(s,a)].
\end{equation}

In practical applications, neural networks are often used to parameterize both the policy and value functions, represented by $\pi_{\theta}$ and $Q_{\phi}$, respectively. These functions are refined through the application of gradient descent methods aimed at reducing the loss functions for both the critic, $\mathcal{L}_{q_{}}(\theta) = \mathbb{E} _ {({s},{a},{s}^{\prime})\sim\mathcal{B}}\left[\left(r(s,a) + \gamma \mathbb{E} _ {s' \sim p, a' \sim \pi}[Q^{\pi}(s',a')]-Q^{\pi} _ {\phi}(s, a)\right)^{2}\right]$, and the actor, $\mathcal{L}_{\pi}(\phi) = -\mathbb{E} _ {s \sim d _ {\pi},a \sim \pi }[Q^{\pi _ {\mathrm{old}}}(s,a)]$. These loss functions are structured based on the principles outlined in \eqref{eq:bellman} and \eqref{eq:policy}.

\subsection{Diffusion Models}
\label{sec:pre_diffusion}
Diffusion models \cite{sohl2015deep, ho2020denoising, song2019generative, song2020score} are highly effective generative tools. They convert data from its original distribution to a Gaussian noise distribution by gradually adding noise and then reconstruct the data by gradually removing this noise through a reverse process. This process is typically described as a continuous Markov chain: the forward process incrementally increases the noise level, while the reverse process involves a conditional generative model trained to predict the optimal reverse transitions at each denoising step. Consequently, the model reverses the diffusion sequence to generate data samples starting from pure noise.

Let us define $p_{\theta}(\boldsymbol{x}_{0}):=\int p_{\theta}(\boldsymbol{x}_{0: T}) \text{d} \boldsymbol{x}_{1: T}$, where $\boldsymbol{x}_{1}, \ldots, \boldsymbol{x}_{T}$ denote latent variables sharing the same dimensionality as the data variable $\boldsymbol{x}_{0} \sim q(\boldsymbol{x}_{0})$, where $q(\boldsymbol{x}_0)$ means original data distribution. In a forward diffusion chain, the noise is incrementally introduced to the data $\boldsymbol{x}_0 \sim q(\boldsymbol{x}_0)$ across $T$ steps, adhering to a predetermined variance sequence denoted by $\beta_t$, described as
\begin{equation}
    q(\boldsymbol{x}_{1:T}|\boldsymbol{x}_0) =  \prod_{t=1}^{T} q(\boldsymbol{x}_t|\boldsymbol{x}_{t-1}), \quad q(\boldsymbol{x}_t|\boldsymbol{x}_{t-1}) = \mathcal{N} (\boldsymbol{x}_t; \sqrt{1 - \beta_t} \boldsymbol{x}_{t-1}, \beta_t).  
\end{equation}
When $T \rightarrow \infty$, $\boldsymbol{x}_T$ distributes as an isotropic Gaussian distribution \cite{kang2024efficient}. The reverse diffusion process of the diffusion model can be represented as
\begin{equation}
\label{eq:cal action}
p_{\theta}(\boldsymbol{x}_{0:T})=p(\boldsymbol{x}_{T})\prod_{t=1}^{T}p_{\theta}(\boldsymbol{x}_{t-1}|\boldsymbol{x}_{t}),\quad p_{\theta}(\boldsymbol{x}_{t-1}|\boldsymbol{x}_{t})=\mathcal{N}(\boldsymbol{x}_{t-1};\boldsymbol{\mu}_{\theta}(\boldsymbol{x}_{t},t),\boldsymbol{\Sigma}_{\theta}(\boldsymbol{x}_{t},t)),
\end{equation}
where $p(\boldsymbol{x}_{T})=\mathcal{N}(\boldsymbol{x}_{T}; \boldsymbol{0}, \boldsymbol{I})$ under the condition that $\prod_{t=1}^{T}(1-\beta_{t}) \approx 0$. 

\section{Method}
\label{sec:method}
In this section, we detail the design of our diffusion actor-critic with entropy regulator (DACER). First, we consider the reverse diffusion process of the diffusion model as a new policy approximator, serving as the policy function in RL. Second, We directly optimize the diffusion policy using gradient descent, whose objective function is to maximize expected Q-values. This feature allows it to be integrated with mainstream RL algorithms that do not require entropy. However, the diffusion policy learned this way produces overly deterministic actions with poor performance. When attempting to integrate the maximization entropy RL framework, we find the entropy of the diffusion policy is difficult to analytically determine. Therefore, we use GMM to approximate the entropy of the diffusion policy, and then learn a parameter $\alpha$ based on it to adjust the exploration level of diffusion policy. 

\subsection{Diffusion Policy Representation}
We use the reverse process of a conditional diffusion model as a parametric policy:
\begin{equation}
\label{eq:diffusion pi}
\pi_{\theta}(\boldsymbol{a}|\boldsymbol{s})=p_{\theta}(\boldsymbol{a}_{0:T}|\boldsymbol{s})=p(\boldsymbol{a}_{T})\prod_{t=1}^{T}p_{\theta}(\boldsymbol{a}_{t-1}|\boldsymbol{a}_{t}, \boldsymbol{s}),
\end{equation}
where $p(\boldsymbol{a}_{T})=\mathcal{N}(0, \boldsymbol{I})$, the end sample of the reverse chain, $\boldsymbol{a}_0$, is the action used for RL evaluation. Generally, $p_{\theta}(\boldsymbol{a}_{t-1}|\boldsymbol{a}_{t}, \boldsymbol{s})$ could be modeled as a Gaussian distribution $\mathcal{N}(\boldsymbol{a}_{t-1};\boldsymbol{\mu}_{\theta}(\boldsymbol{a}_{t}, \boldsymbol{s},t),\boldsymbol{\Sigma}_{\theta}(\boldsymbol{a}_{t},\boldsymbol{s},t))$. We choose to parameterize $\pi_{\theta}(\boldsymbol{a}|\boldsymbol{s})$ like DDPM \cite{ho2020denoising}, which sets $\boldsymbol{\Sigma}_{\theta}(\boldsymbol{a}_{t}, \boldsymbol{s}, t)=\beta_{t}I$ to fixed time-dependent constants, and constructs the mean $\boldsymbol{\mu}_{\theta}$ from a noise prediction model as
\begin{equation}
\boldsymbol{\mu}_\theta(\boldsymbol{a}_t,\boldsymbol{s},t)=\frac{1}{\sqrt{\alpha_t}}\left(\boldsymbol{a}_t-\frac{\beta_t}{\sqrt{1-\bar{\alpha}_t}}\boldsymbol{\epsilon}_\theta(\boldsymbol{a}_t,\boldsymbol{s},t)\right),
\end{equation}
where $\alpha_{t}=1-\beta_{t},\bar{\alpha}_{t}=\prod_{k=1}^{t} \alpha_k$, and $\epsilon_{\theta}$ is a parametric model.

To obtain an action from DDPM, we need to draw samples from $T$ different Gaussian distributions sequentially. The sampling process can be reformulated as
\begin{equation}
\label{eq: diffusion sample}
\boldsymbol{a}_{t-1}=\frac{1}{\sqrt{\alpha_{t}}}\left(\boldsymbol{a}_{t}-\frac{\beta_{t}}{\sqrt{1-\bar{\alpha}_{t}}}\boldsymbol{\epsilon}_{\theta}(\boldsymbol{a}_{t}, \boldsymbol{s}, t)\right)+\sqrt{\beta_{t}}\boldsymbol{\epsilon},
\end{equation}
with the reparametrization trick, where $\boldsymbol{\epsilon} \sim \mathcal{N}(0, \boldsymbol{I})$, $t$ is the reverse timestep from $T$ to $0$, $\boldsymbol{a}_T \sim \mathcal{N}(0,\boldsymbol{I})$.

\subsection{Diffusion Policy Learning}
\label{sec:diffusion policy learning}
In integrating diffusion policy with offline RL, policy improvement relies on minimizing the behavior-cloning term. However, in online RL, without a dataset to imitate, we discarded the behavior-cloning term and the imitation learning framework. In this study, the policy-learning objective is to maximize the expected Q-values of the actions generated by the diffusion network given the state:
\begin{equation}
\label{eq:max_q}
    \max_\theta 
    \mathbb{E}_{\boldsymbol{s}\sim \mathcal{B}, \boldsymbol{a}_0\sim \pi_\theta(\cdot | \boldsymbol{s})}
    \left[Q_\phi(\boldsymbol{s}, \boldsymbol{a}_0)\right].
\end{equation}

Unlike the traditional reverse diffusion process, our study requires recording the gradient of the whole process. The gradient of the Q-value function with respect to the action is backpropagated through the entire diffusion chain.

Policy improvement is introduced above; next, we introduce policy evaluation. The Q-value function is learned through a conventional approach, which involves minimizing the Bellman operator \cite{fujimoto2019off, li2023reinforcement, sutton2018reinforcement} with the double Q-learning trick \cite{van2016deep}. We built two Q-networks $Q_{\phi_1}(\boldsymbol{s}, \boldsymbol{a}), Q_{\phi_2}(\boldsymbol{s}, \boldsymbol{a})$, and target network $Q_{\phi_{1}^{'}}(\boldsymbol{s}, \boldsymbol{a}), Q_{\phi_{2}^{'}}(\boldsymbol{s}, \boldsymbol{a})$. Then we give the objective function of policy evaluation, which is shown as 
\begin{equation}
\min_ {\phi_i} \mathbb{E}_{(\boldsymbol{s},\boldsymbol{a},\boldsymbol{s}')\sim\mathcal{B}}\left[\left( \left(r(\boldsymbol{s},\boldsymbol{a})+\gamma \min_ {i=1,2} Q_{{\phi_{i}^{'}}}(\boldsymbol{s}',\boldsymbol{a}')\right)-Q_{\phi_i}(\boldsymbol{s},\boldsymbol{a})\right)^2\right],
\end{equation}
where $\boldsymbol{a}^{'}$ is obtained by inputting the $\boldsymbol{s}^{'}$ into the diffusion policy, $\mathcal{B}$ means replay buffer. Building on this, we employ the tricks in DSAC \cite{duan2023dsac, duan2021distributional} to mitigate the problem of Q-value overestimation.

The diffusion policy we construct can be directly combined with mainstream RL algorithms that do not require policy entropy. However, training with the above diffusion policy learning method suffers from overly deterministic policy actions, resulting in poor performance of the final diffusion policy. In the next section, we will propose entropy estimation to solve this problem and obtain diffusion policy with SOTA performance.

\subsection{Diffusion Policy with Entropy}
\label{sec:diffusion entropy}
The diffusion policy's distribution lacks an analytic expression, so we cannot directly determine its entropy. However, in the same state, we can use multiple samples to obtain a series of actions. By fitting these action points, we can estimate the action distribution corresponding to the state.

In this paper, we use Gaussian mixture model (GMM) to fit the policy distribution. The GMM forms a complex probability density function by combining multiple Gaussian distributions, which can be represented as
\begin{equation}
\label{eq:GMM}
\hat{f}(\boldsymbol{a})=\sum_{k=1}^Kw_k \cdot \mathcal{N}(\boldsymbol{a}|\boldsymbol{\mu}_k,\boldsymbol{\Sigma}_k),
\end{equation}
where $K$ is the number of Gaussian distributions, and $w_k$ is the mixing weight of the $k$-th component, satisfying $\sum_{k=1}^Kw_k=1, w_k\geq0$. $\boldsymbol{\mu}_k$, $\boldsymbol{\Sigma}_k$ are the mean and covariance matrices of the $k$-th Gaussian distribution, respectively.

For each state,  we use a diffusion policy to sample $N$ actions, $\boldsymbol{a}^1, \boldsymbol{a}^2, \dots, \boldsymbol{a}^N \in \mathcal{A}$. The Expectation-Maximization algorithm is then used to estimate the parameters of the GMM. In the expectation step, the posterior probability that each data point $\boldsymbol{a}^i$ belongs to each component $k$ is computed, denoted as
\begin{equation}
\label{eq:GMM params pre}
\gamma(\boldsymbol{z}^{i}_{k}) = \frac{w_k \cdot \mathcal{N}(\boldsymbol{a}^i | \boldsymbol{\mu}_k, \boldsymbol{\Sigma}_k)}{\sum_{j=1}^K w_j \cdot \mathcal{N}(\boldsymbol{a}^i | \boldsymbol{\mu}_j, \boldsymbol{\Sigma}_j)},
\end{equation}
where $\gamma(\boldsymbol{z}^{i}_{k})$ denotes that under the current parameter estimates, the observed data $\boldsymbol{a}^i$ come from the $k$-th component of the probability. In the maximization step, the results of the Eq. \eqref{eq:GMM params pre} calculations are used to update the parameters and mixing weights for each component:
\begin{equation}
w_k=\frac1N\sum_{i=1}^N\gamma(\boldsymbol{z}^{i}_{k}), \boldsymbol{\mu}_k=\frac{\sum_{i=1}^N\gamma(\boldsymbol{z}^{i}_{k})\cdot \boldsymbol{a}^i}{\sum_{i=1}^N\gamma(\boldsymbol{z}^{i}_{k})}, \boldsymbol{\Sigma}_k=\frac{\sum_{i=1}^N\gamma(\boldsymbol{z}^{i}_{k})(\boldsymbol{a}^i-\boldsymbol{\mu}_k)(\boldsymbol{a}^i-\boldsymbol{\mu}_k)^\text{T}}{\sum_{i=1}^N\gamma(\boldsymbol{z}^{i}_{k})}.
\end{equation}
Iterative optimization continues until parameter convergence. Based on our experimental experience in the MuJoCo environments, a general setting of $K=3$ provides a better fit to the action distribution.

According to Eq. \eqref{eq:GMM}, we can estimate the entropy of the action distribution corresponding to the state by \cite{huber2008entropy}
\begin{equation}
\mathcal{H}_{\boldsymbol{s}} \approx -\sum_{k=1}^K w_k\log w_k + \sum_{k=1}^Kw_k \cdot \frac12\log\left((2\pi e)^d|\boldsymbol{\Sigma}_k|\right),
\end{equation}
where $d$ is the dimension of action. Then, the mean of the entropy of the actions associated with the chosen batch of states is used as the estimated entropy $\hat{\mathcal{H}}$ of the diffusion policy.

Similar to maximizing entropy RL, we learn a parameter $\alpha$ based on the estimated entropy. We update this parameter using
\begin{equation}
\label{eq:alpha_learning}
\alpha\leftarrow \alpha-\beta_{\alpha}[\hat{\mathcal{H}} -\overline{\mathcal{H}}],
\end{equation}
where $\overline{\mathcal{H}}$ is target entropy. Finally, we use $\boldsymbol{a} = \boldsymbol{a} + \lambda \alpha \cdot \mathcal{N}(0, \boldsymbol{I})$ to adjust the diffusion policy entropy during training, where $\lambda$ is a hyperparameter and $\boldsymbol{a}$ is the output of diffusion policy. Additionally, no noise is added during the evaluation phase. We summarize our implementation in Algorithm \ref{alg:diffusion policy}.


\begin{algorithm}[!htb]
  \caption{Diffusion Actor-Critic with Entropy Regulator for Online RL}
  \label{alg:diffusion policy}
  \begin{algorithmic}
  \STATE Input: $\lambda$, $\theta$, $\phi_1$, $\phi_2$, $\phi_{1}^{'}$, $\phi_{2}^{'}$, $\alpha$, $\beta_q$, $\beta_{\alpha}$, $\beta_{\pi}$, and $\rho$
  
  \FOR{each iteration}
  
  \FOR{each sampling step}
  \STATE Sample $\boldsymbol{a} \sim \pi_{\theta}(\cdot | \boldsymbol{s})$ by Eq. \eqref{eq:diffusion pi}
  \STATE Add noise $\boldsymbol{a} = \boldsymbol{a} + \lambda \alpha \cdot\mathcal{N}(0, \boldsymbol{I})$
  \STATE Get reward $\boldsymbol{r}$ and new state $\boldsymbol{s'}$
  \STATE Store a batch of samples $\boldsymbol{(s,a,r,s')}$ in replay buffer $\mathcal{B}$ 
  \ENDFOR
  
  \FOR{each update step}
  \STATE Sample data from $\mathcal{B}$
  \STATE Update critic networks using $\phi_i \leftarrow \phi_i - \beta_{q}\nabla_{\phi_i} \mathcal{L}_{q}(\phi_i) $ for $i= \{ 1, 2 \} $
  \STATE Update diffusion policy network using $\theta \leftarrow \theta - \beta_{\pi}\nabla_{\theta} \mathcal{L}_{\pi}(\theta) $

  \STATE \textbf{if} step $\bmod$ 10000 $==$ 0 \textbf{then}
  \STATE \quad Estimate the entropy of diffusion policy $\hat{\mathcal{H}} = \mathbb{E}_{\boldsymbol{s} \sim \mathcal{B}} \left[ \mathcal{H}_{\boldsymbol{s}} \right]$
    \STATE \quad Update $\alpha$ using Eq. \eqref{eq:alpha_learning}
  \STATE Update target networks using $\phi_{i}^{\prime}=\rho\phi_{i}^{\prime}+(1-\rho)\phi_{i}\mathrm{~for~}i=\{1,2\}$
  \ENDFOR
  \ENDFOR
\end{algorithmic}
\end{algorithm}

\section{Experiments}
\label{sec:experiment}
We evaluate the performance of our method in some control tasks of RL within MuJoCo \cite{2012MuJoCo}. The benchmark tasks utilized in this study are depicted in Fig. \ref{fig:mujoco}, including Humanoid-v3, Ant-v3, HalfCheetah-v3, Walker2d-v3, InvertedDoublePendulum-v3, Hopper-v3, Pusher-v2, and Swimmer-v3. Moreover, we conducted experiments in a multi-goal task to demonstrate the excellent representational and exploratory capabilities of our diffusion policy. We also provide ablation studies on the critical components for better understanding. All baseline algorithms are available in GOPS \cite{wang2023gops}, an open-source RL solver developed with PyTorch.

\paragraph{Baselines.} Our algorithm is compared and evaluated against the six well-known model-free algorithms. These include DDPG \cite{silver2014deterministic}, TD3 \cite{fujimoto2018addressing}, PPO \cite{schulman2017proximal}, SAC \cite{haarnoja2018soft}, DSAC \cite{duan2023dsac, duan2021distributional}, and TRPO \cite{schulman2015trust}. These baselines have been extensively tested and applied in a series of demanding domains.

\paragraph{Experimental details.} To ensure a fair comparison, we incorporated the diffusion policy as a policy approximation function within GOPS and implemented DACER with JAX, which improves training speed by 4-5 times compared to PyTorch while maintaining consistent performance. All algorithms and tasks use the same three-layer MLP neural network with GeLU \cite{hendrycks2016gaussian} or Mish \cite{misra2019mish} activation functions, the latter used only for the noise prediction network in the diffusion policy.  Initially, we encode timestep $t$ into 16 dimensions using sinusoidal embedding \cite{vaswani2017attention}, then merge this encoded result with the state $\boldsymbol{s}$ and action $\boldsymbol{a}_t$ during the current denoising step, and input it into the prediction noise network to generate the output. The impact of the reverse diffusion step size, $T$, on the experimental results will be examined in the ablation experiments. $T$ is set to 20 for all experiments eventually. The Adam \cite{kingma2014adam} optimization method is employed for all parameter updates. In this paper, the total training step size for all experiments is set at 1.5 million, with the results of all experiments averaged over five random seeds. The CPU used for the experiment is the AMD Ryzen Threadripper 3960X 24-Core Processor, and the GPU is NVIDIA GeForce RTX 3090Ti. Taking Humanoid-v3 as an example, the time taken to train 1.5 million in the JAX framework is 6 hours. More detailed hyperparameters are provided in Appendix \ref{app:hyper} due to space limits.

\paragraph{Evaluation Protocol.} In this paper, we use the same assessment metrics as DSAC. For each seed, the metric is derived by averaging the highest return values observed during the final 10\% of iteration steps in each run, with evaluations conducted every 15,000 iterations. Each assessment result is the average of ten episodes. The results from the five seeds are then aggregated to calculate the mean and standard deviation. Additionally, the training curves in Fig. \ref{fig:benchmark} provide insights into the stability of the training process.

\subsection{Comparative Evaluation}
\label{sec:compare Mujoco}
\begin{figure*}[t!]
    \centering
    \begin{subfigure}[b]{0.24\textwidth}
        \includegraphics[width=\textwidth]{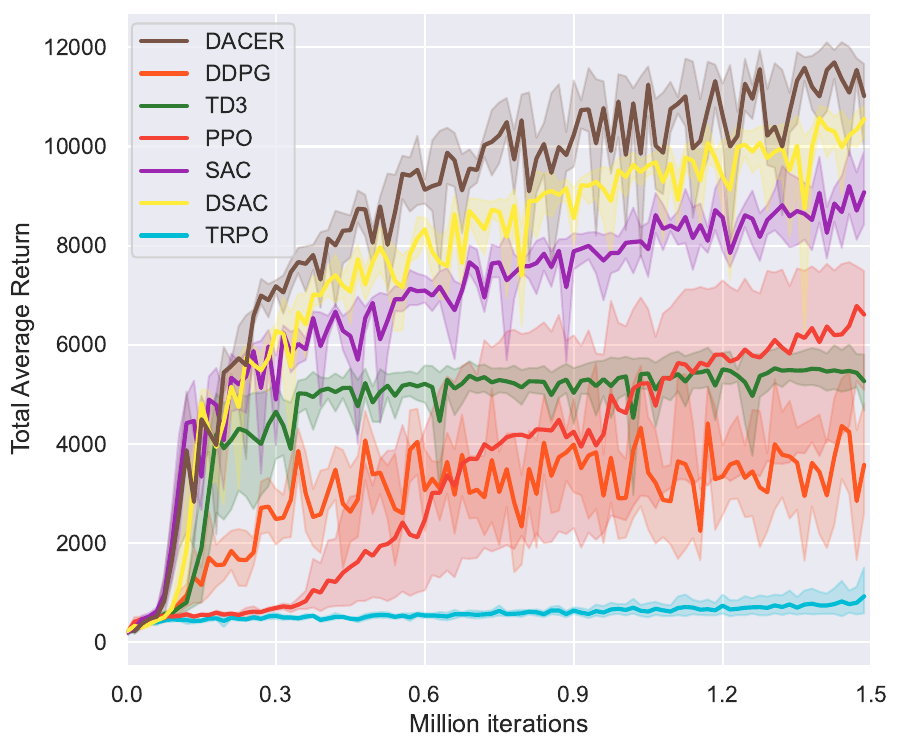}
        \caption{Humanoid-v3}
        \label{fig:tar-humanoid}
    \end{subfigure}
    \begin{subfigure}[b]{0.24\textwidth}
        \includegraphics[width=\textwidth]{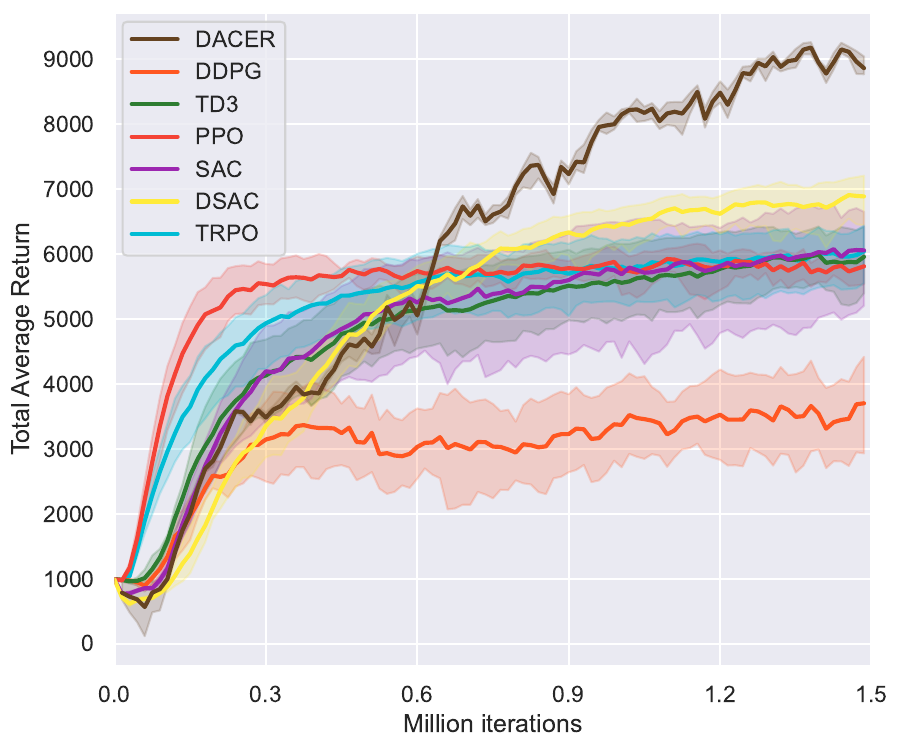}
        \caption{Ant-v3}
        \label{fig:tar-ant}
    \end{subfigure}
    \begin{subfigure}[b]{0.24\textwidth}
        \includegraphics[width=\textwidth]{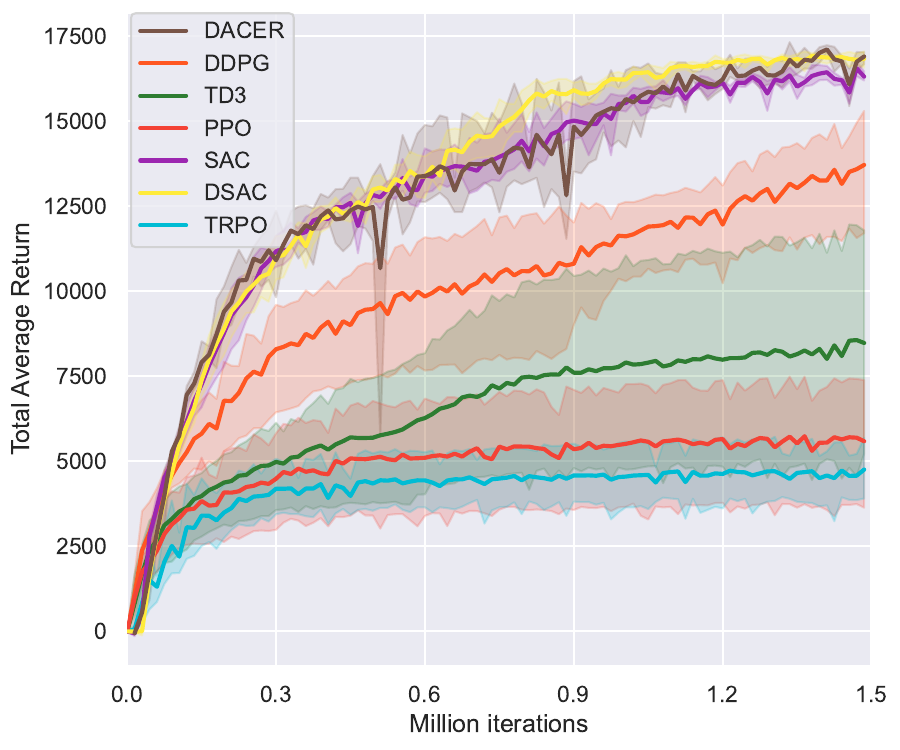}
        \caption{HalfCheetah-v3}
        \label{fig:tar-half}
    \end{subfigure}
    \begin{subfigure}[b]{0.24\textwidth}
        \includegraphics[width=\textwidth]{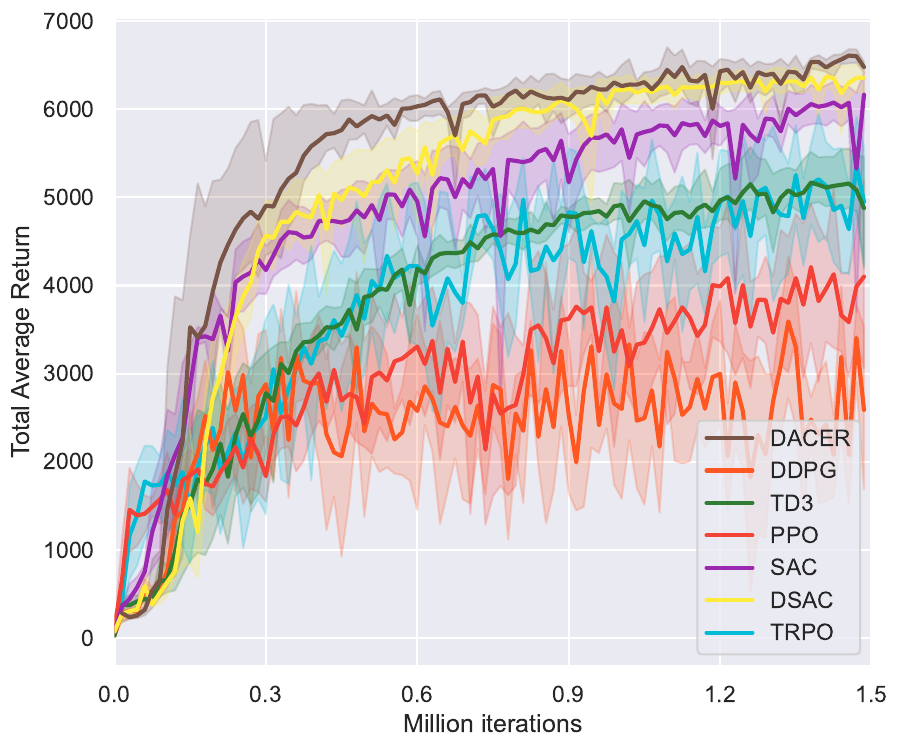}
        \caption{Walker2d-v3}
        \label{fig:tar-walker}
    \end{subfigure}
    
    \begin{subfigure}[b]{0.24\textwidth}
        \includegraphics[width=\textwidth]{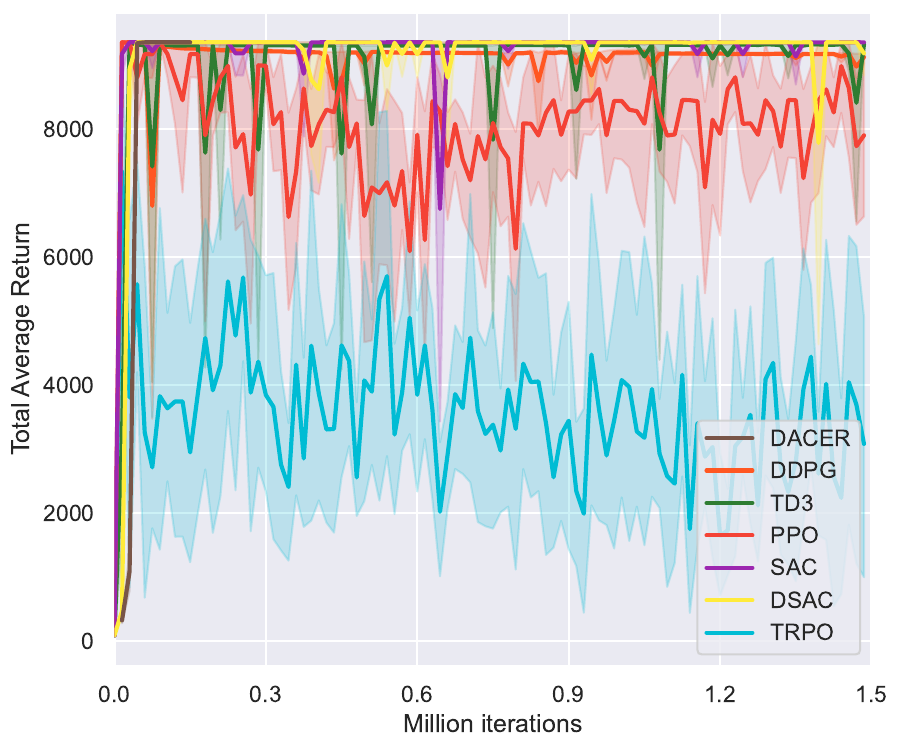}
        \caption{Inverted2Pendulum-v3}
        \label{fig:tar-inverted}
    \end{subfigure}
    \begin{subfigure}[b]{0.24\textwidth}
        \includegraphics[width=\textwidth]{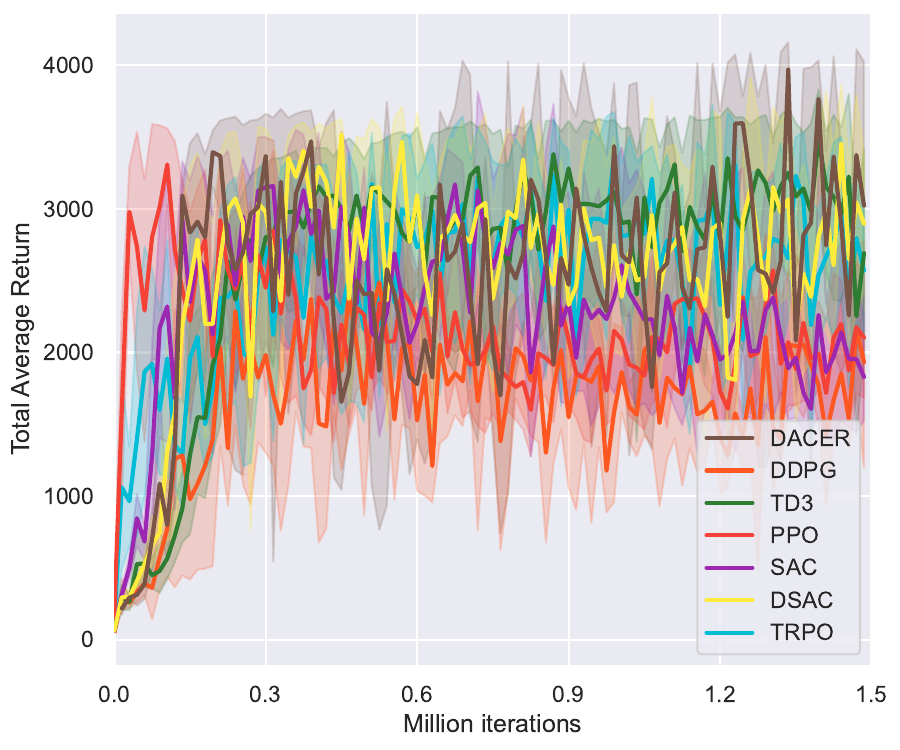}
        \caption{Hopper-v3}
        \label{fig:tar-hopper}
    \end{subfigure}
    \begin{subfigure}[b]{0.24\textwidth}
        \includegraphics[width=\textwidth]{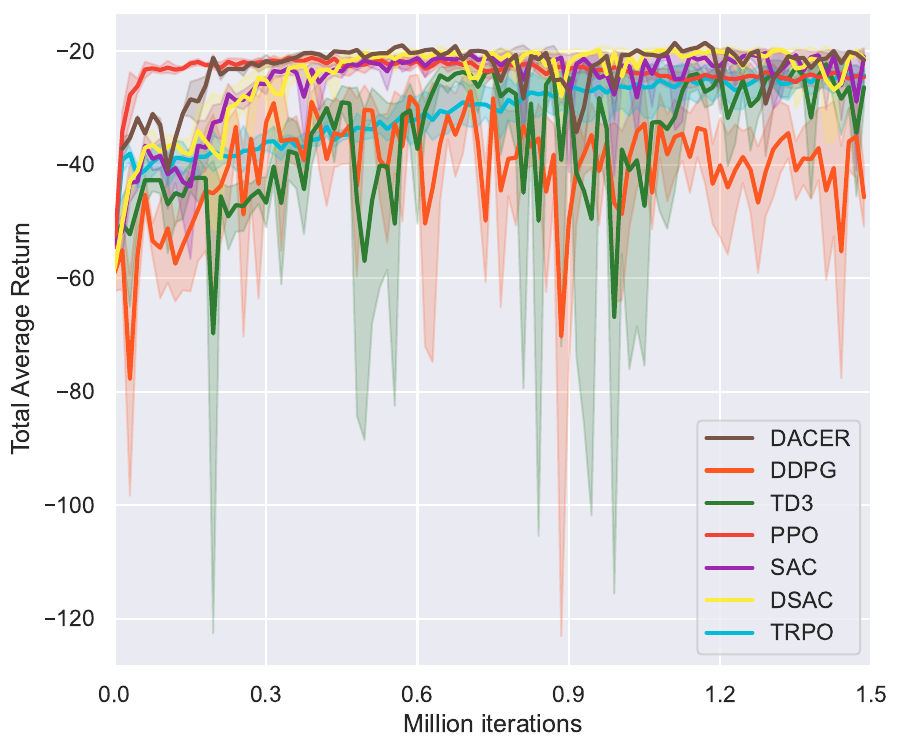}
        \caption{Pusher-v2}
        \label{fig:tar-pusher}
    \end{subfigure}
    \begin{subfigure}[b]{0.24\textwidth}
        \includegraphics[width=\textwidth]{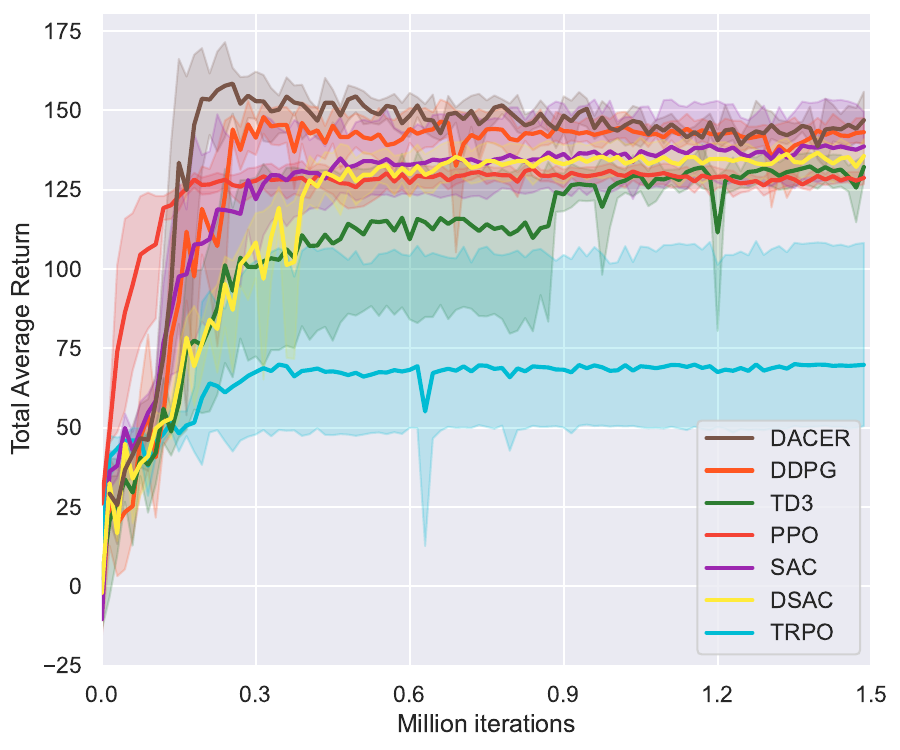}
        \caption{Swimmer-v3}
        \label{fig:tar-Swimmer}
    \end{subfigure}

\caption{\textbf{Training curves on benchmarks.} The solid lines represent the mean, while the shaded regions indicate the 95\% confidence interval over five runs. The iteration of PPO and TRPO is measured by the number of network updates.}
\label{fig:benchmark}
\end{figure*}

Each algorithm was subjected to five distinct tests, utilizing a variety of consistent random seeds to ensure robustness in the results. Fig. \ref{fig:benchmark} and Table \ref{tab:benchmark} display the learning curves and performance strategies, respectively. Our comprehensive findings reveal that across all evaluated tasks, the DACER algorithm consistently matched or surpassed the performance of all competing benchmark algorithms. Specifically, in the Humanoid-v3 scenario, our algorithm demonstrated enhancements of 124.7\%, 111.1\%, 73.1\%, 27.3\%, 9.8\%, and 1131.9\% over DDPG, TD3, PPO, SAC, DSAC, and TRPO, respectively.

\begin{table*}[!htb]
 \centering
\captionsetup{justification=centering,labelsep=newline,font={small}}
    \caption{\textcolor{black}{\textbf{Average final return.} Computed as the mean of the highest return values observed in the final 10\% of iteration steps per run, with an evaluation interval of 15,000 iterations. The maximum value for each task is bolded. $\pm$ corresponds to standard deviation over five runs.}}

	\label{tab:benchmark}
   \resizebox{\textwidth}{!}{ 
\begin{tabular}{c c c c c c c c}
  \toprule
  Task & DACER & DSAC & SAC & TD3 & DDPG & TRPO & PPO \\ \hline
  Humanoid-v3 & \textbf{11888} $\pm$ \textbf{244}& 10829 $\pm$ 243 & 9335 $\pm$ 695 & 5631 $\pm$ 435 & 5291 $\pm$ 662 & 965 $\pm$ 555 & 6869 $\pm$ 1563 \\
  Ant-v3 & \textbf{9108} $\pm$ \textbf{103}& 7086 $\pm$ 261& 6427 $\pm$ 804 & 6184 $\pm$ 486 & 4549 $\pm$ 788 & 6203 $\pm$ 578 & 6156 $\pm$ 185 \\
  Halfcheetah-v3 & \textbf{17177} $\pm$ \textbf{176}& 17025 $\pm$ 157 & 16573 $\pm$ 224 & 8632 $\pm$ 4041 & 13970 $\pm$ 2083 & 4785 $\pm$ 967 & 5789 $\pm$ 2200 \\
  Walker2d-v3 & \textbf{6701} $\pm$ \textbf{62}& 6424 $\pm$ 147 & 6200 $\pm$ 263 & 5237 $\pm$ 335 & 4095 $\pm$ 68 & 5502 $\pm$ 593 & 4831 $\pm$ 637 \\
  Inverteddoublependulum-v3 & \textbf{9360} $\pm$ \textbf{0} & \textbf{9360} $\pm$ \textbf{0} & \textbf{9360} $\pm$ \textbf{0} & 9347 $\pm$ 15 & 9183 $\pm$ 9 & 6259 $\pm$ 2065 & 9356 $\pm$ 2 \\
  Hopper-v3 & \textbf{4104} $\pm$ \textbf{49}&  3660$\pm$533& 2483 $\pm$ 943 & 3569 $\pm$ 455 & 2644 $\pm$ 659 & 3474 $\pm$ 400 & 2647 $\pm$ 482 \\
  Pusher-v2 & \textbf{-19} $\pm$ \textbf{1} &  \textbf{-19} $\pm$ \textbf{1} & -20 $\pm$ 0 & -21 $\pm$ 1 & -30 $\pm$ 6 & -23 $\pm$ 2 & -23 $\pm$ 1 \\
  Swimmer-v3 & \textbf{152} $\pm$ \textbf{7} &  138$\pm$6& 140 $\pm$ 14 & 134 $\pm$ 5 & 146 $\pm$ 4 & 70 $\pm$ 38& 130 $\pm$ 2 \\
  \bottomrule
\end{tabular}
}
\end{table*}

\subsection{Policy Representation Experiment}

\begin{figure*}[t!]
    \centering
    \begin{subfigure}[b]{0.223\textwidth}
        \includegraphics[width=\textwidth]{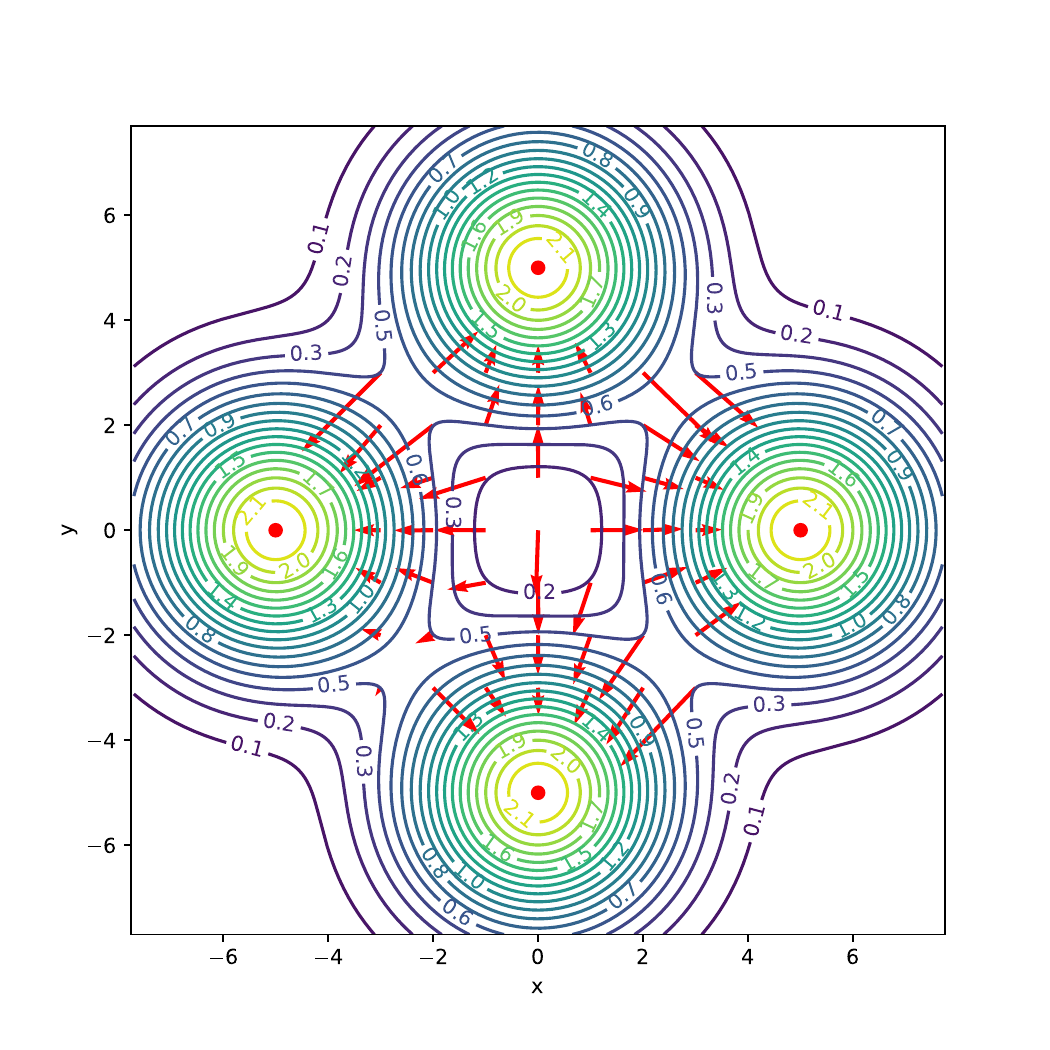}
        \label{fig:dac-action}
    \end{subfigure}
    \hfill
    \begin{subfigure}[b]{0.222\textwidth}
        \includegraphics[width=\textwidth]{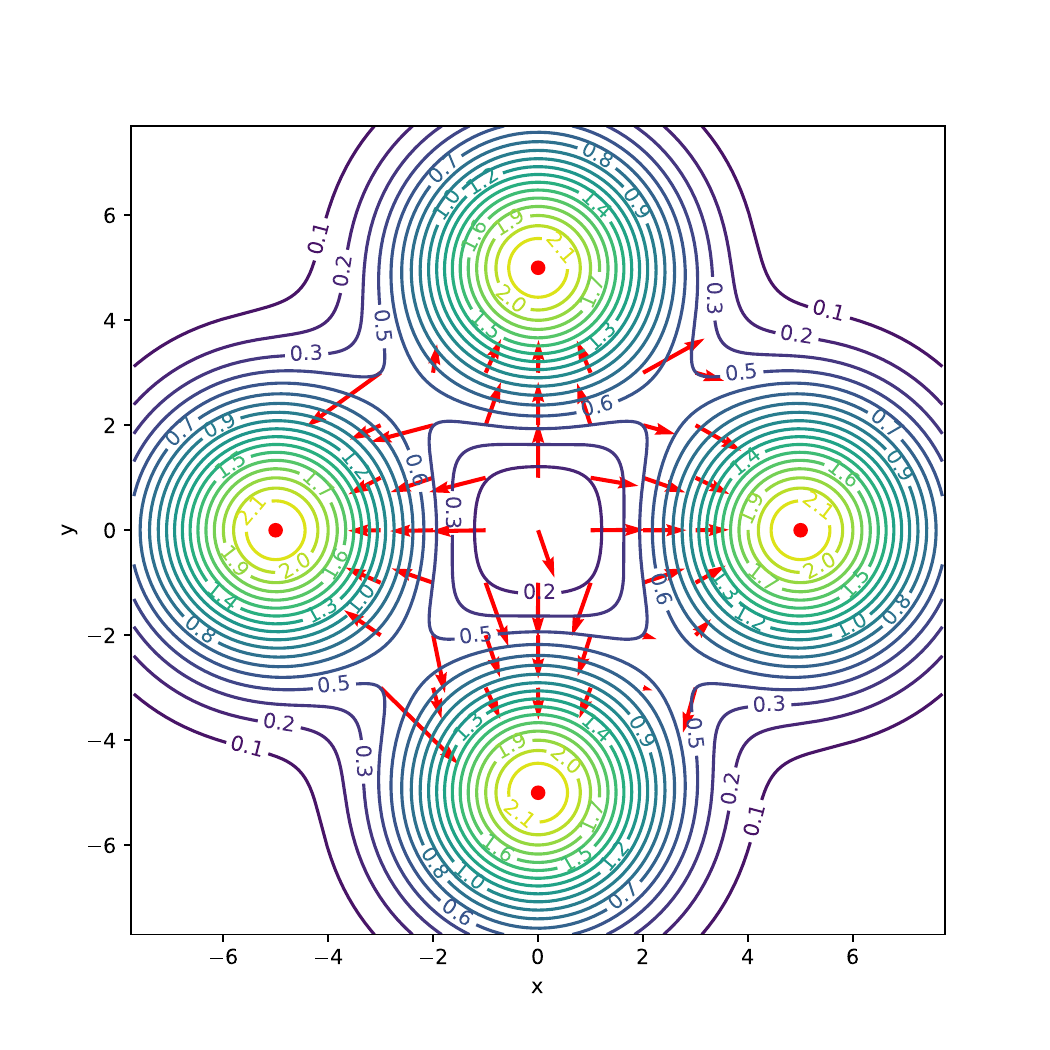}
        \label{fig:dsac-action}
    \end{subfigure}
    \hfill
    \begin{subfigure}[b]{0.23\textwidth}
        \includegraphics[width=\textwidth]{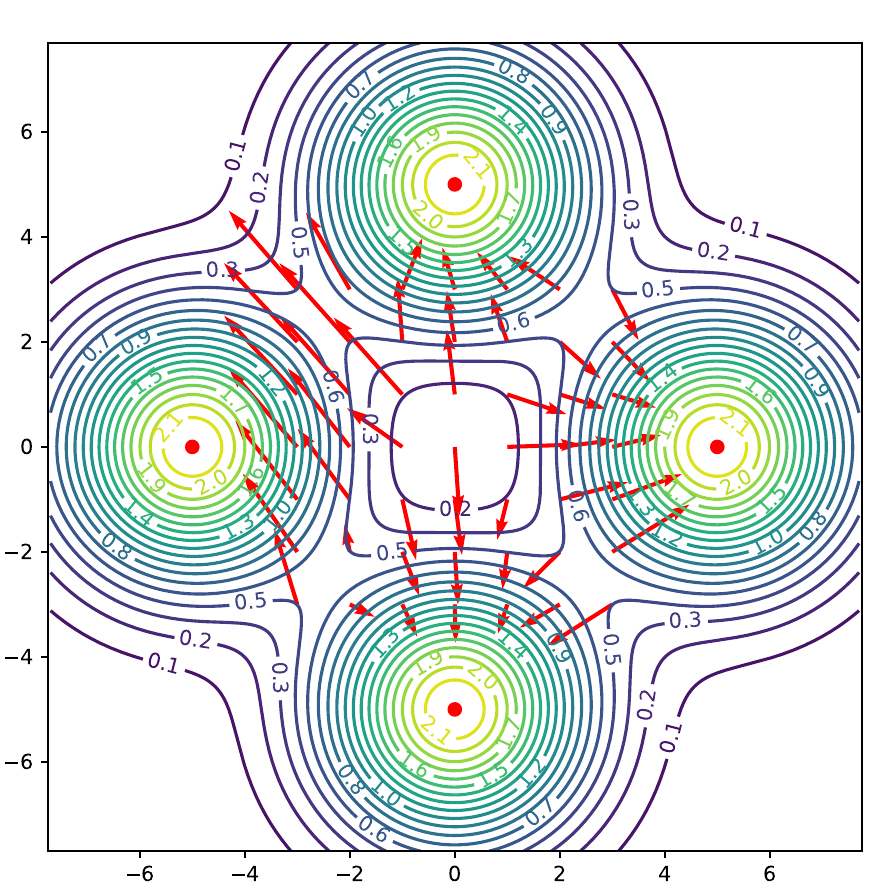}
        \label{fig:td3-action}
    \end{subfigure}
    \hfill
    \begin{subfigure}[b]{0.23\textwidth}
        \includegraphics[width=\textwidth]{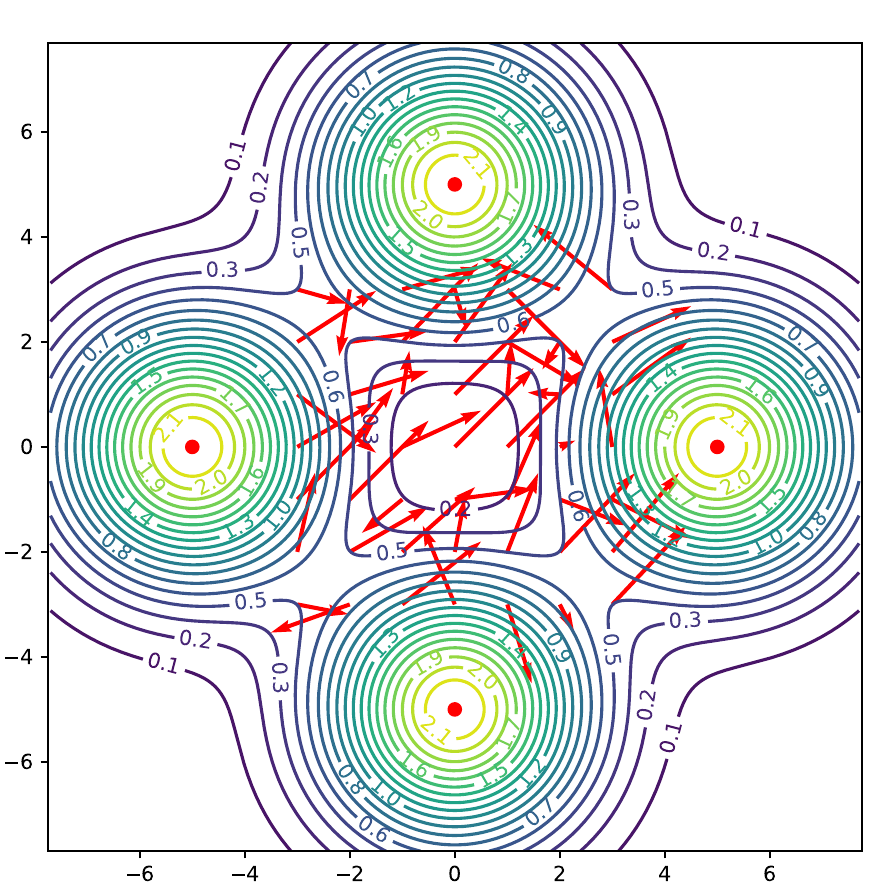}
        \label{fig:ppo-action}
    \end{subfigure}

    \begin{subfigure}[b]{0.25\textwidth}
        \includegraphics[width=\textwidth]{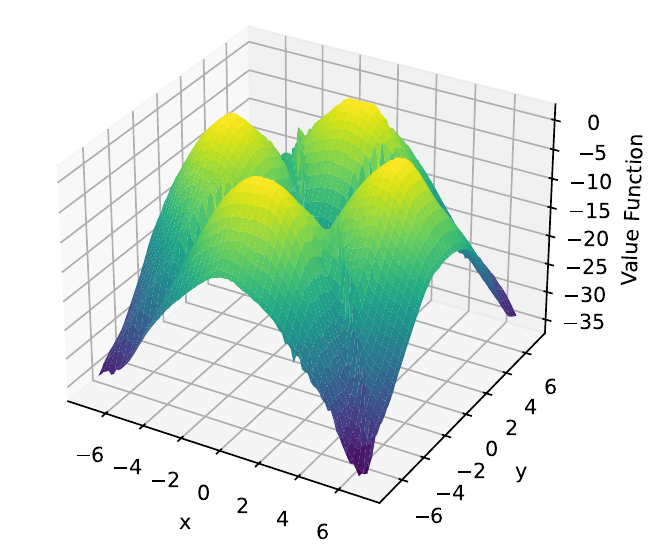}
        \caption{DACER}
        \label{fig:dac-q}
    \end{subfigure}
    \hfill
    \begin{subfigure}[b]{0.23\textwidth}
        \includegraphics[width=\textwidth]{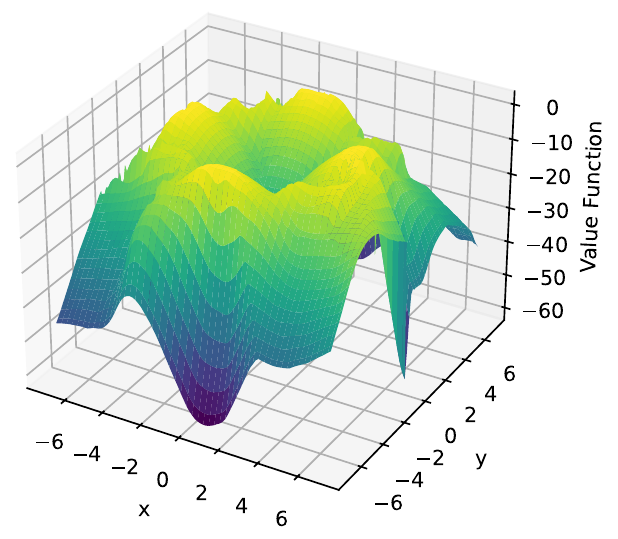}
        \caption{DSAC}
        \label{fig:dsac-q}
    \end{subfigure}
    \hfill
    \begin{subfigure}[b]{0.23\textwidth}
        \includegraphics[width=\textwidth]{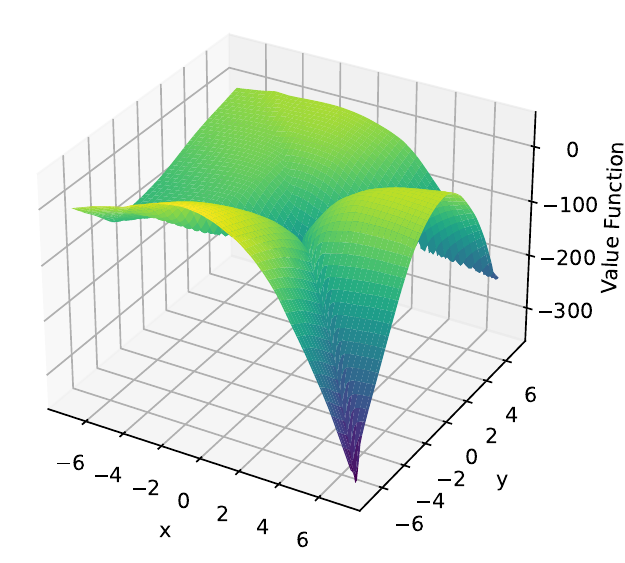}
        \caption{TD3}
        \label{fig:td3-q}
    \end{subfigure}
    \hfill
    \begin{subfigure}[b]{0.23\textwidth}
        \includegraphics[width=\textwidth]{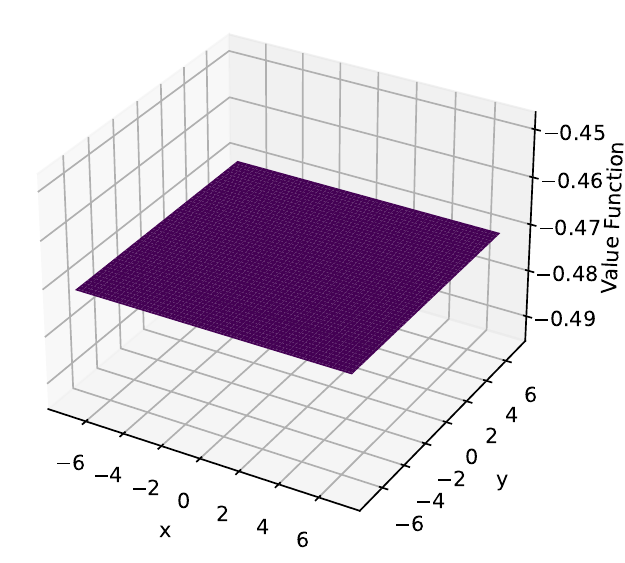}
        \caption{PPO}
        \label{fig:ppo-q}
    \end{subfigure}

    \caption{\textbf{Policy representation comparison of different policies on a multimodal environment.} The first row exhibits the policy distribution. The length of the red arrowheads denotes the size of the action vector, and the direction of the red arrowheads denotes the direction of actions. The second row shows the value function of each state point.}
    \label{fig:comparison-multi-goal}
\end{figure*}

In this section, we conduct an experiment to confirm the representation capability of the diffusion policy. We use an environment called "Multi-goal" \cite{haarnoja2017reinforcement}, as shown in Fig. \ref{fig:comparison-multi-goal}, where the $x$-axis and $y$-axis represent 2D states. In this setup, the agent is represented as a 2D point mass situated on a $7*7$ plane. The objective for the agent is to navigate towards one of four symmetrically positioned points: $(0, 5)$, $(0, -5)$, $(5, 0)$, and $(-5, 0)$. Since the goal positions are symmetrically distributed at the four points, a policy with strong representational capacity should enable the Q-function to learn the four symmetric peaks across the entire state space. This result reflects the policy's capacity for exploration in understanding the environment.

We compare the performance of DACER with DSAC, TD3, and PPO, as shown in Fig. \ref{fig:comparison-multi-goal}. The results show that DACER's actions are likely to point to the nearest peak in different states. DACER's value function curve shows four symmetrical peaks, aligning with the previous analysis. Compared to DSAC, our method learns a better policy representation, mainly due to using a diffusion policy instead of an MLP. In contrast, TD3 and PPO generate more random actions with poorer policy representation, lacking the symmetrical peaks in their value function curves. Overall, our method demonstrates superior representational capability.

\begin{figure}[ht!]
  \centering
    \includegraphics[width=1\textwidth]{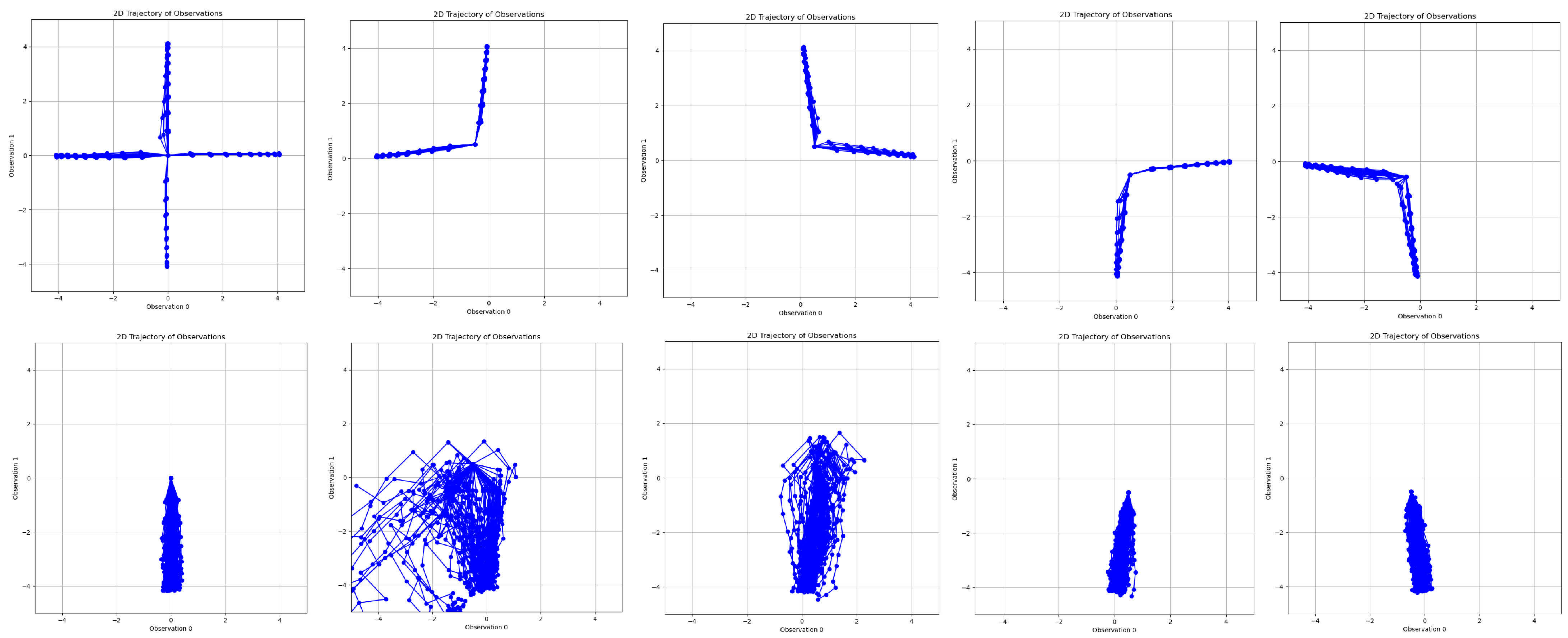} 
  \caption{\textbf{Multi-goal multimodal experiments.} We selected 5 points that require multimodal policies: (0, 0), (-0.5, 0.5), (0.5, 0.5), (0.5, -0.5), (-0.5, -0.5), and sampled 100 trajectories for each point. The top row shows the experimental results of DACER, another shows the experimental results of DSAC.
}
  \label{fig: multimodal}
\end{figure}

To demonstrate the powerful multimodality of DACER, we select five points requiring multimodal policies: (0.5, 0.5), (0.5, -0.5), (-0.5, -0.5), (-0.5, 0.5), and (0, 0). For each point, we sampled 100 trajectories. The trajectories are plotted in Fig. \ref{fig: multimodal}. The results show that compared with DSAC, DACER exhibits strong multimodality. This also explains why only the Q-function of DACER can learn the nearly perfectly symmetrical four peaks.

\subsection{Ablation Study}
\label{sec:ablation}
In this section, we analyze why DACER outperforms all other baseline algorithms on MuJoCo tasks. We conduct ablation experiments to investigate the impact of the following three aspects on the performance of the diffusion policy: 1) whether adding Gaussian noise to the final output action of the diffusion policy; 2) whether the standard deviation of the added Gaussian noise can be adaptively adjusted by estimated entropy; 3) different reverse diffusion step size $T$.

\begin{figure*}[ht!]
\centering
  \begin{subfigure}{0.3\linewidth} 
    \centering
    \includegraphics[width=\textwidth]{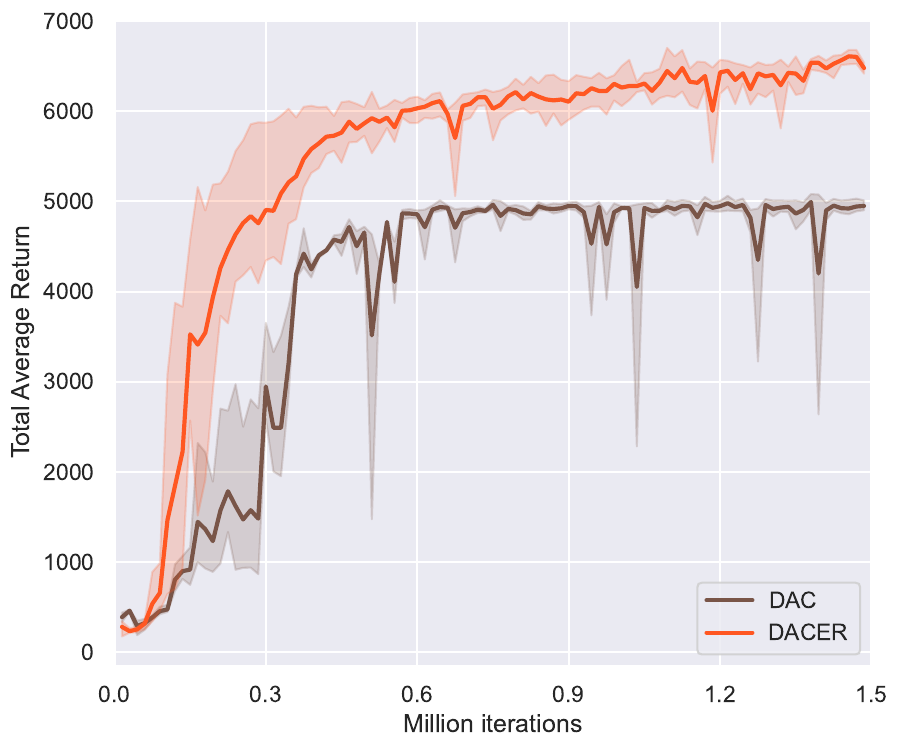} 
    \label{fig:different qloss}
    \caption{\centering{Ablation for the entropy regulator mechanism.}}
  \end{subfigure}
  \hfill
  \begin{subfigure}{0.3\linewidth}
    \centering
    \includegraphics[width=\textwidth]{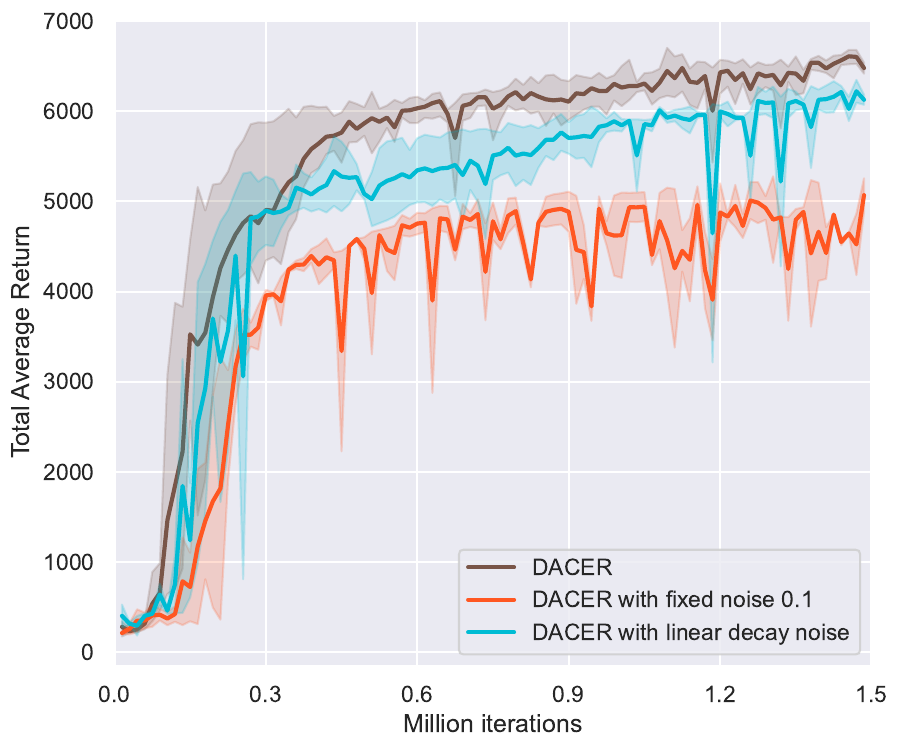} 
    \label{fig:different noise}
    \caption{\centering{Ablation for the noise factor modulation mechanism.}}
  \end{subfigure}
  \hfill
  \begin{subfigure}{0.3\linewidth}
    \centering
    \includegraphics[width=\textwidth]{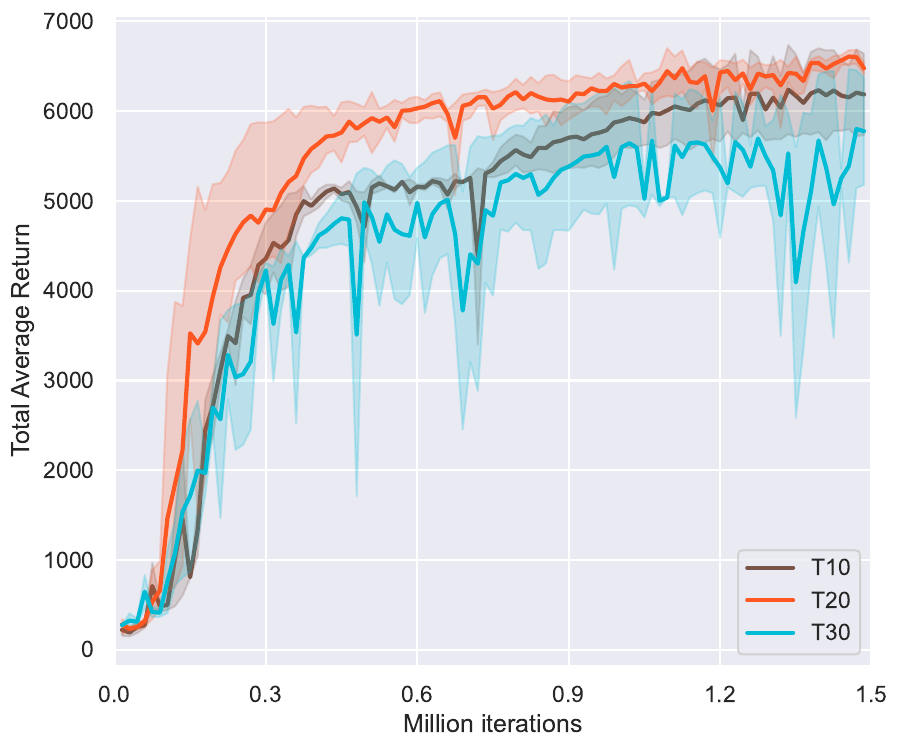} 
    \label{fig:diffusion steps}
    \caption{\centering{Ablation for the different diffusion steps.}}
  \end{subfigure}
  
\caption{\textbf{Ablation experiment curves.} (a) DAC stands for not using the entropy regulator. DACER's performance on Walker2d-v3 is far better than DAC. (b) Adaptive tuning of the noise factor based on the estimated entropy achieved the best performance compared to fixing the noise factor or using the adaptive tuning method with initial, end values followed by a linear decay method. (c) The best performance was achieved with diffusion steps equal to 20, in addition to the instability of the training process when equal to 30.} 
\label{fig: DACER Ablation}
\end{figure*}

\paragraph{Only Q-learning.} In section \ref{sec:diffusion policy learning}, we propose a method using the reverse diffusion process as a policy approximator, which can be combined with the non-maximizing entropy RL algorithm. However, the diffusion policy trained without entropy exhibits poor exploratory properties, leading to suboptimal performance. Using Walker2d-v3 as an example, we compared the training curves of this method with the DACER algorithm, as shown in Fig. \ref{fig: DACER Ablation}(a).

\paragraph{Fixed and linear decay noise factor.} 
In order to verify that using the estimated entropy to adaptively adjust the noise factor plays an important role in the final performance, we conducted the following two experiments in the Walker2d-v3 task: 1) Fixed noise factor to 0.1; 2) The noise factor starts from 0.27 and linearly decreases to 0.1 during the training process. These two values were chosen because the starting and ending noise factor for adaptive tuning in this setting is about in this range. As shown in Fig. \ref{fig: DACER Ablation}(b), our method of adaptively adjusting the noise factor based on the estimated entropy achieves the best performance.


\paragraph{Diffusion steps.} We further examined the performance of the diffusion policy as the number of diffusion timesteps $T$ varied. We used the Walker2d-v3 task to plot training curves for $T=10, 20, $ and $30$, as shown in Fig. \ref{fig: DACER Ablation}(c). Experimental results indicate that a larger number of diffusion steps does not necessarily lead to better performance. Excessive diffusion steps can cause gradient explosion, significantly reducing the performance of diffusion policy. After balancing performance and computational efficiency, we selected 20 diffusion steps for all experiments.

\section{Conclusion}
\label{sec:conclusion}
In this study, we propose the diffusion actor-critic with entropy regulator (DACER) algorithm, a novel RL method designed to overcome the limitations of traditional RL methods that use diagonal Gaussian distributions for policy parameterization. By utilizing the inverse process of the diffusion model, DACER effectively handles multimodal distributions, enabling the creation of more complex policies and improving policy performance. A significant challenge arises from the lack of analytical expressions to determine the entropy of a diffusion strategy. To address this, we employ GMM to estimate entropy, thereby facilitating the learning of a key parameter, $\alpha$, which adjusts the exploration-exploitation balance by regulating the noise variance in the action output. Empirical tests on the MuJoCo benchmark and a multimodal task show the superior performance of DACER.

\section{Acknowledgements}
This study is supported by National Key R\&D Program of China with 2022YFB2502901, and Tsinghua University Initiative Scientific Research Program. 

\newpage
\bibliographystyle{plain}
\bibliography{ref}

\newpage
\appendix

\section{Environmental Details}
\subsection{Experimental Environment Introduction}

\begin{figure}[ht!]
  \centering
    \includegraphics[width=0.75\textwidth]{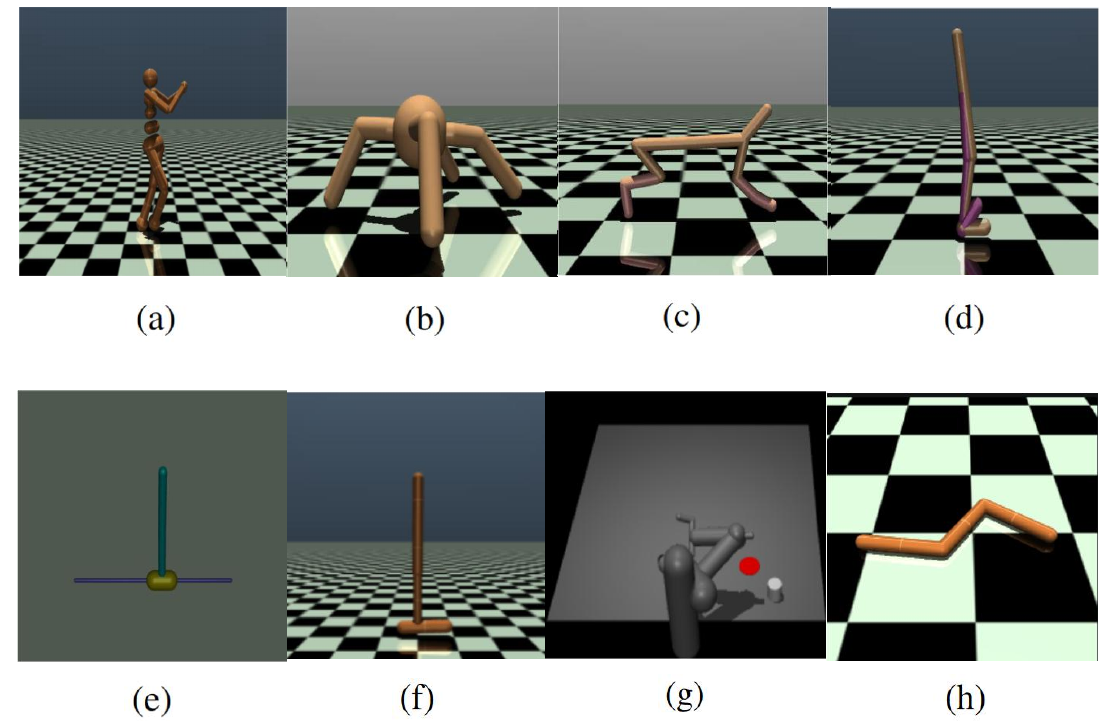} 
  \caption{\textbf{Simulation tasks.} (a) Humanoid-v3:$(s \times a) \in \mathbb{R}^{376} \times \mathbb{R}^{17}$. 
    (b) Ant-v3: $(s \times a) \in \mathbb{R}^{111} \times \mathbb{R}^{8}$. (c) HalfCheetah-v3 : $(s \times a) \in \mathbb{R}^{17} \times \mathbb{R}^{6}$. (d) Walker2d-v3: $(s \times a) \in \mathbb{R}^{17} \times \mathbb{R}^{6}$. (e) InvertedDoublePendulum-v3: $(s \times a) \in \mathbb{R}^{6} \times \mathbb{R}^{1}$. (f) Hopper-v3: $(s \times a) \in \mathbb{R}^{11} \times \mathbb{R}^{3}$. (g) Pusher-v2: $(s \times a) \in \mathbb{R}^{23} \times \mathbb{R}^{7}$. (h) Swimmer-v3: $(s \times a) \in \mathbb{R}^{8} \times \mathbb{R}^{2}$.}
  \label{fig:mujoco}
\end{figure}

The benchmark tasks utilized in this study are depicted in Fig. \ref{fig:mujoco}, including Humanoid-v3, Ant-v3, HalfCheetah-v3, Walker2d-v3, InvertedDoublePendulum-v3, Hopper-v3, Pusher-v2, and Swimmer-v3. 
\label{app:environment introduction}

\subsection{Training Details on MuJoCo tasks}
\label{app:hyper}
\begin{table}[!htp]
\centering
\captionsetup{justification=centering,labelsep=newline,font={small,sc}}
\caption{Detailed hyperparameters.}
\label{tab:baseline hyper}
\begin{tabular}{lc}
\toprule
Hyperparameters & Value \\
\hline
\emph{Shared} & \\
\quad Replay buffer capacity & 1000000 \\
\quad Buffer warm-up size & 30000 \\
\quad Batch size & 256 \\
\quad Initial alpha $\alpha$ & 0.272 \\
\quad Action bound & $[-1, 1]$ \\
\quad Hidden layers in critic network & [256, 256, 256] \\
\quad Hidden layers in actor network & [256, 256, 256] \\
\quad Activation in critic network & GeLU \\
\quad Activation in actor network & GeLU \\
\quad Optimizer &  Adam ($\beta_{1}=0.9, \beta_{2}=0.999$)\\
\quad Actor learning rate & $1{\rm{e-}}4 $\\
\quad Critic learning rate & $1{\rm{e-}}4 $\\
\quad Discount factor ($\gamma$) & 0.99\\
\quad Policy update interval & 2\\
\quad Target smoothing coefficient ($\rho$) & 0.005\\
\quad Reward scale & 0.2\\
\hline
\emph{Maximum-entropy framework} &\\ 
\quad  Learning rate of $\alpha$ &  $3{\rm{e-}}4 $ \\
\quad  Expected entropy ($\overline{\mathcal{H}}$) &  $\overline{\mathcal{H}}=-{\rm{dim}}(\mathcal{A})$ \\
\hline
\emph{Deterministic policy} &\\ 
\quad Exploration noise&  $\epsilon \sim \mathcal{N}(0,0.1^2)$\\
\hline
\emph{Off-policy} &\\ 
\quad Replay buffer size & $1\times10^6$\\
\quad Sample batch size &  20 \\
\hline
\emph{On-policy} &\\ 
\quad Sample batch size &  2000 \\
\quad Replay batch size &  2000 \\
\bottomrule
\end{tabular}
\end{table}

\begin{table}[ht!]
\centering
\captionsetup{justification=centering,labelsep=newline,font={small,sc}}
\caption{Algorithm hyperparameter}
\label{tab: DAC hyperparameter of Mujoco}
\begin{tabular}{@{}lc@{}}
\toprule
\textbf{Parameter} & \textbf{Setting} \\
\midrule
Replay buffer capacity & 1000000 \\
Buffer warm-up size & 30000 \\
Buffer warm-up size (Humanoid, HalfCheetah) & 200000 \\ 
Batch size & 256 \\
Discount $\gamma$ & 0.99 \\
Initial alpha $\alpha$ & 0.27 \\
Target network soft-update rate $\rho$ & 0.005 \\
Network update times per iteration & 1 \\
Action bound & $[-1, 1]$ \\
Reward scale & 0.2\\
Hidden layers in noise prediction network & [256, 256, 256] \\
Hidden layers in noise prediction network (Humanoid)& [512, 512, 512] \\
Hidden layers in critic network & [256, 256, 256] \\
Activations in critic network & GeLU \\
Activations in actor network & Mish \\
Policy act distribution & TanhGauss \\
Policy min log std & -20 \\
Policy max log std & 0.5 \\
Policy delay update & 2 \\
Number of Gaussian distributions for mixing & 3 \\
Number of action samples in entropy estimation & 200 \\
Alpha delay update & 10000 \\
Noise scale $\lambda$ & 0.1 \\
Noise scale $\lambda$ (Humanoid, HalfCheetah) & 0.15 \\
Optimizer & Adam \\
Actor learning rate & $1 \cdot 10^{-4}$ \\
Critic learning rate & $1 \cdot 10^{-4}$ \\
Alpha learning rate & $3 \cdot 10^{-2}$ \\
Target entropy & $-0.9 \cdot {\rm{dim}}(\mathcal{A})$\\
\bottomrule
\end{tabular}
\end{table}

Mujoco \cite{2012MuJoCo} is a simulation engine primarily designed for research in RL and robotics. It provides a versatile, physics-based platform for developing and testing various RL algorithms. Core features of Mujoco include a highly efficient physics engine, realistic modeling of dynamic systems, and support for complex articulated robots. Currently, it is one of the most recognized benchmark environments for RL and continuous control.

The hyperparameters of all baseline algorithms are shown in Table \ref{tab:baseline hyper}. Moreover, the hyperparameters of the DACER in the MuJoCo task are shown in Table \ref{tab: DAC hyperparameter of Mujoco}.

\section{Limitation and Future Work}
\label{app:limitation}
In this study, we propose using GMM to estimate the entropy of the diffusion policy and, based on this estimate, learn a parameter $\alpha$ to balance exploration and exploitation.
However, the process of estimating entropy requires a large number of samples and takes a long time (about 40 ms). Therefore, we estimate the entropy of diffusion policy every 10,000 iterations to reduce the impact on training time. But, this approach prevents perfect integration with maximizing entropy RL. In future work, we will avoid using batch-size data to estimate entropy and find a balance between estimation accuracy and computational efficiency so as to better combine our method with maximizing entropy RL.

\section{Positive and Negative Social Impact}
\label{app:social impact}
In this paper, we propose DACER, an online RL algorithm that uses the reverse diffusion process as a policy approximator. Diffusion policy has powerful multimodal representation capabilities, making it widely applicable in complex environments such as automated manufacturing, autonomous driving, and industrial control. However, DACER could also enhance the exploratory capabilities and operational efficiency of military AI, potentially posing threats to citizen privacy and security.

\newpage
\section*{NeurIPS Paper Checklist}

\begin{enumerate}

\item {\bf Claims}
    \item[] Question: Do the main claims made in the abstract and introduction accurately reflect the paper's contributions and scope?
    \item[] Answer: \answerYes{} 
    \item[] Justification:  In both the abstract and the introduction we give a formulation of the contribution points. These include considering the reverse diffusion process as a novel policy approximator, proposing a Gaussian mixture model method to estimate the entropy of the diffusion policy and learning a parameter $\alpha$ for the regulation of the policy exploration level, and open-sourcing the code. Experimental results in eight MuJoCo environments as well as one multimodal environment demonstrate the good performance of our method and the strong characterization capability of the diffusion policy.


\item {\bf Limitations}
    \item[] Question: Does the paper discuss the limitations of the work performed by the authors?
    \item[] Answer: \answerYes{} 
    \item[] Justification: In Section \ref{app:limitation} of the appendix, we acknowledge the limitations of our methodology, estimating entropy requires a considerable amount of time, making it difficult to perfectly integrate this method with maximum entropy RL. However, in future work, we will improve the estimation efficiency by reducing the number of states used for estimation.

\item {\bf Theory Assumptions and Proofs}
    \item[] Question: For each theoretical result, does the paper provide the full set of assumptions and a complete (and correct) proof?
    \item[] Answer: \answerNA{} 
    \item[] Justification: 

    \item {\bf Experimental Result Reproducibility}
    \item[] Question: Does the paper fully disclose all the information needed to reproduce the main experimental results of the paper to the extent that it affects the main claims and/or conclusions of the paper (regardless of whether the code and data are provided or not)?
    \item[] Answer: \answerYes{} 
    \item[] Justification: First, in Section \ref{sec:diffusion entropy},  we provide pseudo-code for the algorithm \ref{alg:diffusion policy} of diffusion actor-critic with entropy regulator (DACER). Second, in Section \ref{sec:experiment}, we provide details of the experimental environment setup. Finally, among the two tables in the Appendix \ref{app:hyper}, we give details of the hyperparameter configuration.

\item {\bf Open access to data and code}
    \item[] Question: Does the paper provide open access to the data and code, with sufficient instructions to faithfully reproduce the main experimental results, as described in supplemental material?
    \item[] Answer: \answerYes{} 
    \item[] Justification: All the reproductions of the experimental results of the baseline algorithms can be obtained by running the general optimal control problem solver (GOPS). And GOPS can be searched on github. We provide the PyTorch code for DACER as a function approximator in GOPS. Besides, we have implemented the DACER algorithm in JAX. 
    
    Regarding the DACER algorithm, we give an implementation of all the core parts, but the complete training code we will open source after the review.


\item {\bf Experimental Setting/Details}
    \item[] Question: Does the paper specify all the training and test details (e.g., data splits, hyperparameters, how they were chosen, type of optimizer, etc.) necessary to understand the results?
    \item[] Answer: \answerYes{} 
    \item[] Justification: In Appendix \ref{app:hyper}, we give the design of all hyperparameters, including optimizer selection, learning rate, neural network configuration, and so on.

\item {\bf Experiment Statistical Significance}
    \item[] Question: Does the paper report error bars suitably and correctly defined or other appropriate information about the statistical significance of the experiments?
    \item[] Answer: \answerYes{}{} 
    \item[] Justification: All results of this experiment came from in 5 random seeds. In Section \ref{sec:experiment}, we mention the specific way of evaluation and the fact that all results are presented as mean ± standard deviation. Regarding the training curves, the solid lines represent the mean, while the shaded regions indicate the 95\% confidence interval over five runs.

\item {\bf Experiments Compute Resources}
    \item[] Question: For each experiment, does the paper provide sufficient information on the computer resources (type of compute workers, memory, time of execution) needed to reproduce the experiments?
    \item[] Answer: \answerYes{} 
    \item[] Justification: In Section \ref{sec:experiment}, we provide the CPU and GPU models used for training, which are Xeon(R) Platinum 8352V and NVIDIA GeForce RTX 4090, respectively. We take the example of training 1.5 million Humanoid-v3, and the time required to train it in the JAX framework is about 7 hours.
    
\item {\bf Code Of Ethics}
    \item[] Question: Does the research conducted in the paper conform, in every respect, with the NeurIPS Code of Ethics \url{https://neurips.cc/public/EthicsGuidelines}?
    \item[] Answer: \answerYes{} 
    \item[] Justification: We made sure the code was anonymous.

\item {\bf Broader Impacts}
    \item[] Question: Does the paper discuss both potential positive societal impacts and negative societal impacts of the work performed?
    \item[] Answer: \answerYes{} 
    \item[] Justification: In Appendix \ref{app:social impact}, we discuss the potential positive and negative social impacts of our work.
    
\item {\bf Safeguards}
    \item[] Question: Does the paper describe safeguards that have been put in place for responsible release of data or models that have a high risk for misuse (e.g., pretrained language models, image generators, or scraped datasets)?
    \item[] Answer: \answerNA{} 
    \item[] Justification: 

\item {\bf Licenses for existing assets}
    \item[] Question: Are the creators or original owners of assets (e.g., code, data, models), used in the paper, properly credited and are the license and terms of use explicitly mentioned and properly respected?
    \item[] Answer: \answerYes{} 
    \item[] Justification: We used the GOPS solver for training and cited the corresponding paper. This open-source code library is licensed under the Apache-2.0 license. Copyright © 2022 Intelligent Driving Laboratory (iDLab). All rights reserved.

\item {\bf New Assets}
    \item[] Question: Are new assets introduced in the paper well documented and is the documentation provided alongside the assets?
    \item[] Answer: \answerNA{} 
    \item[] Justification:

\item {\bf Crowdsourcing and Research with Human Subjects}
    \item[] Question: For crowdsourcing experiments and research with human subjects, does the paper include the full text of instructions given to participants and screenshots, if applicable, as well as details about compensation (if any)? 
    \item[] Answer: \answerNA{} 
    \item[] Justification: 

\item {\bf Institutional Review Board (IRB) Approvals or Equivalent for Research with Human Subjects}
    \item[] Question: Does the paper describe potential risks incurred by study participants, whether such risks were disclosed to the subjects, and whether Institutional Review Board (IRB) approvals (or an equivalent approval/review based on the requirements of your country or institution) were obtained?
    \item[] Answer: \answerNA{} 
    \item[] Justification:

\end{enumerate}

\end{document}